\documentclass[runningheads]{llncs}

\usepackage{eccv}

\usepackage{eccvabbrv}

\usepackage{graphicx}
\usepackage{booktabs}

\usepackage[accsupp]{axessibility}  % Improves PDF readability for those with disabilities.

\usepackage{ifthen}
\usepackage{color}
\usepackage{bigstrut}
\usepackage{multirow}
\usepackage{mathtools}
\usepackage[normalem]{ulem}
\usepackage{tabularx}
\usepackage[dvipsnames]{xcolor}
\usepackage{algorithm}
\usepackage{algpseudocode}
\usepackage{rotating}
\usepackage{footnote}

\usepackage[pagebackref,breaklinks,colorlinks,citecolor=eccvblue]{hyperref}
\usepackage{hyperref}

\usepackage{orcidlink}

\begin{document}

\title{Self-training Room Layout Estimation via Geometry-aware Ray-casting} 

\titlerunning{Self-training Room Layout Estimation via Ray-casting}

% \author{Bolivar Solarte$^{1}$, Chin-Hsuan Wu$^{1}$, Jin Cheng Jhang$^{1}$, Jonathan Lee$^{1}$
% \\
% Yi-Hsuan Tsai$^{2}$, Min Sun$^{13}$}
\author{Bolivar Solarte\inst{1, 2}\orcidlink{0000-0003-3518-755X} \and
Chin-Hsuan Wu\inst{1}\orcidlink{0000-0003-1547-0825} \and 
Jin-Cheng Jhang\inst{1}\orcidlink{0009-0006-9777-4951} \and Jonathan Lee\inst{1}\orcidlink{0009-0000-8923-5129}
\and Yi-Hsuan Tsai\inst{2}\orcidlink{0000-0002-6191-0134}
\and Min Sun\inst{1}\orcidlink{0000-0001-9598-8178}
}

\authorrunning{B.~Solarte et al.}

% Springer Heidelberg, Tiergartenstr.~17, 69121 Heidelberg, Germany
% \institute{$^{1}$National Tsing Hua University, $^{2}$Google,
% $^{3}$Amazon}

\institute{National Tsing Hua University, Taiwan \and Industrial Technology Research Institute ITRI, Taiwan \and Google
\\ \href{https://enriquesolarte.github.io/ray-casting-mlc/}{enriquesolarte.github.io/ray-casting-mlc}
}
% \href{https://enriquesolarte.github.io/ray-casting-mlc/}{enriquesolarte.github.io/ray-casting-mlc}} 
% \email{enrique.solarte.pardo@gmail.com}, 
% \email{sunmin@ee.nthu.edu.tw}}
% \email{chinhsuanwu@gapp.nthu.edu.tw}
% Define each member color
\definecolor{red}{rgb}{0.9,0.1,0}
\definecolor{gray}{rgb}{0.8,0.8,0.8}
\definecolor{blue}{rgb}{0.4,0.4,0.9}
\definecolor{green}{rgb}{0, 0.4, 0}
\definecolor{orange}{rgb}{1, 0.5, 0}
\definecolor{slateblue}{rgb}{0.7,0.35,0.9}
\definecolor{mahogany}{rgb}{0.75, 0.25, 0.0}
\definecolor{purple}{rgb}{0.6, 0, 0.6}
\definecolor{goldenrod}{rgb}{0.85, 0.65, 0.13}
\newbool{revising}
\setbool{revising}{false}
\ifbool{revising}
{
    \newcommand{\todo}[1]{{\color{red}#1}}
    \newcommand{\kike}[1]{\textcolor{blue}{#1}}
     \newcommand{\kikecomment}[1]
     {\textcolor{green}{[kike: #1]}}
    \newcommand{\justin}[1]{\textcolor{green}{#1}}
    \newcommand{\justincomment}[1]{\textcolor{green}{[justin: #1]}}
    \newcommand{\frank}[1]{\textcolor{slateblue}{#1}}
    \newcommand{\jonathan}[1]{\textcolor{mahogany}{#1}}
    \newcommand{\dennis}[1]{\textcolor{orange}{#1}}
    \newcommand{\minsun}[1]{\textcolor{magenta}{#1}}
    \newcommand{\idea}[1]{\textcolor{red}{[Idea]:#1}}
} {
    \newcommand{\kike}[1]{#1}
    \newcommand{\justin}[1]{{#1}}
    \newcommand{\frank}[1]{{#1}}
    \newcommand{\jonathan}[1]{{#1}}
    \newcommand{\dennis}[1]{{#1}}
    \newcommand{\minsun}[1]{{#1}}
    \newcommand{\todo}[1]{{}}
    \newcommand{\TODO}[1]{{}}
    \newcommand{\idea}[1]{{}}
}

\maketitle
\begin{abstract}

\kike{In this paper, we introduce a novel geometry-aware self-training framework for room layout estimation models on unseen scenes with unlabeled data.}
\kike{Our approach utilizes a ray-casting formulation to aggregate multiple estimates from different viewing positions, enabling the computation of reliable pseudo-labels for self-training. In particular, our ray-casting approach enforces multi-view consistency along all ray directions and prioritizes spatial proximity to the camera view for geometry reasoning.}
% \kike{Our approach employs a ray-casting formulation to aggregate multiple estimates for computing geometry-aware pseudo-labels by enforcing multi-view consistency as well as reasoning on their proximity to the camera view.}
\kike{As a result, our geometry-aware pseudo-labels effectively handle complex room geometries and occluded walls without relying on assumptions such as Manhattan World or planar room walls.}
\kike{Evaluation on publicly available datasets, including synthetic and real-world scenarios, demonstrates significant improvements in current state-of-the-art layout models without using any human annotation. 
% Code and datasets are available at 
% \href{https://enriquesolarte.github.io/ray-casting-mlc/}{enriquesolarte.github.io/ray-casting-mlc}
}
% \kike{In particular, our approach uses multiple estimates from a pre-trained model to create pseudo-labels via a novel multi-cycle ray-casting process that aggregates multiple noisy estimates along multiple ray directions and locations in an unseen scene.}
% \kike{By constraining this data aggregation to be consistent along multiple views and sampling to the closest layout measurements along each ray direction, our solution is capable of reasoning occluded and outsider regions directly from noisy.}
% \kike{We present MLC++, a geometry-aware self-training solution for room layout estimation using only unlabeled multi-view panoramic images as input.}
% \kike{Our approach leverages a ray-casting mechanism to aggregate multiple layout estimations into a geometry-aware pseudo-label, i.e., a pseudo-label that considers occluded and outsider regions directly from noisy geometry estimates. As a result, our pseudo-labels can effectively leverage unseen scenes with different environmental conditions, complex room geometries, and different architectural styles without any label annotation or geometry constraint.}
% \kike{To enhance the effectiveness of our ray-casting pseudo-labels, we introduce a weighted loss function that prioritizes pseudo-labels' geometry on the farthest walls from the camera view. This stems from the empirical evidence indicating that accurately estimating the farthest geometries from a single view is challenging, suggesting they offer the most informative cues in a multi-view setting.}
\end{abstract}
\section{Introduction}
\label{sec:intro}
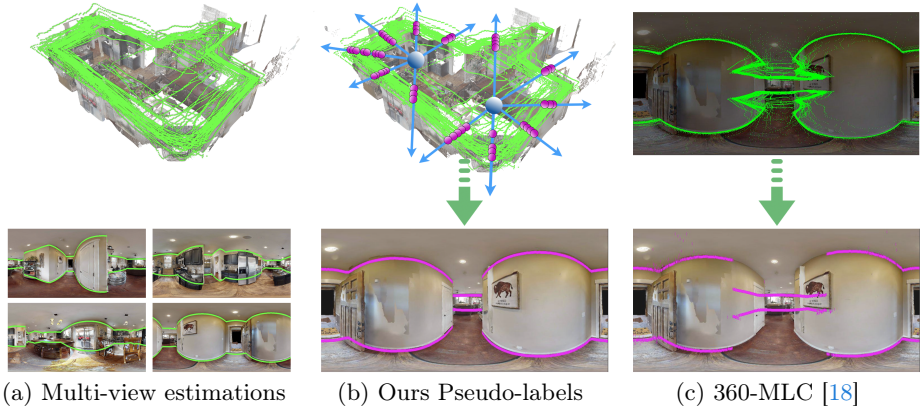
\begin{figure}[h!]
\centering
  \includegraphics[ width=\linewidth]
    {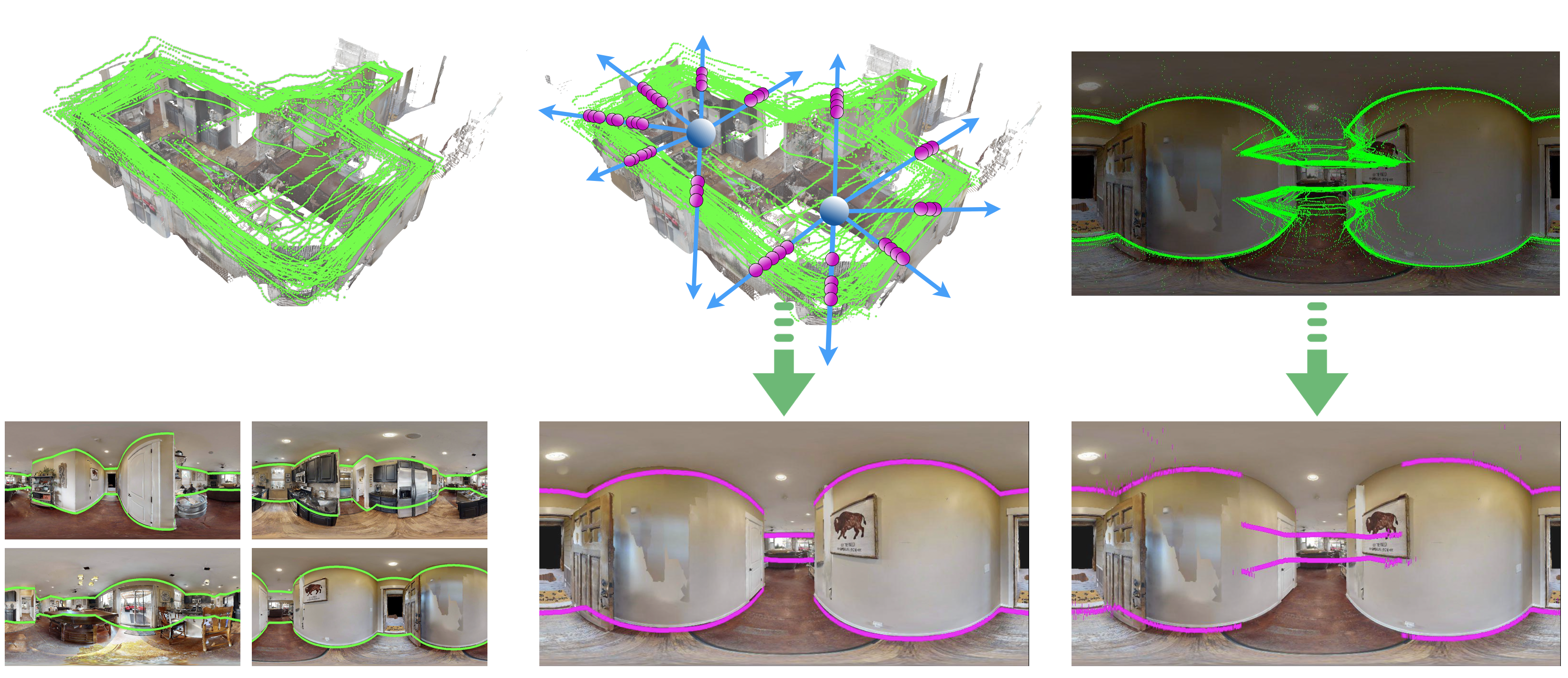}

    \begin{picture}(0, 0)
    
    \put(-175,3){(a) Multi-view estimations}
    
    \put(-48,3){(b) Ours Pseudo-labels}
    
    \put(80,3){(c) 360-MLC~\cite{360_mlc}}
    
    \end{picture}
    % \vspace{-2mm}
    \caption{By leveraging multiple estimates from a pre-trained model as presented in panel (a), Our solution leverages a ray-casting data aggregation process to estimate geometry-aware pseudo-labels for self-training, as depicted in panel (b), i.e., pseudo-labels that encompass a comprehensive representation of the room geometry. In comparison with previous solutions, as presented in (c), where multiple estimations are processed on the image domain without geometry reasoning, our approach excels in defining better pseudo-labels, especially for occluded geometries, highlighting the significance of our contribution.}
    \label{fig_teaser}
    \vspace{-8mm}    
\end{figure}

\kike{While significant progress has been made in room layout estimation, current state-of-the-art solutions predominantly rely on supervised frameworks, utilizing either monocular panoramic images~\cite{horizonnet, hohonet, led2net, lgtnet} or direct geometry sensors like depth cameras or LiDAR~\cite{mp3d, tang2023mvdiffusion}.}
\kike{However, this reliance presents a significant challenge for real-world applications due to variations in geometry complexity and scene conditions, thereby making data collection and manual labeling particularly cumbersome.}
% This reliance presents a significant challenge due to the diverse complexity of indoor environments, making data collection and manual labeling particularly cumbersome.}
% \kike{On the other hand, in practical applications such as robot navigation~\cite{huang2023visual} and 3D reconstruction~\cite{360_dfpe}, where multiple views per scene are commonly available, a solution that directly exploits these views may significantly impact the room layout estimation task.}

\kike{A practical solution for self-training a geometry-based model in unseen environments is by exploiting the multi-view consistency from multiple noisy estimations~\cite{Li_2023_CVPR, monodepth2}. However, applying multi-view consistency for room layout estimation has been poorly explored in the literature.}
% \kike{While leveraging multiple views for geometry perception tasks has been extensively explored in the literature~\cite{Li_2023_CVPR, monodepth2}, its application to room layout estimation has been noticeably limited. 
\kike{For instance, recent approaches in multi-view layout estimation~\cite{graph_covis, gprnet, CoVisPose} particularly rely on ground truth annotations to define important concepts such as wall occlusion and wall match correspondences.  Other solutions avoid partial dependency on label annotation by leveraging a semi-supervised approach~\cite{sslayout}. To the best of our knowledge, only the recent self-training approach, 360-MLC~\cite{360_mlc}, is capable of exploiting multi-view layout consistency (MLC) without human label annotations. Nevertheless, 360-MLC lacks any geometry reasoning and treats all layout estimates from every view equally, leading to noisy pseudo labels, especially for occluded regions. See~\cref{fig_teaser}-(c).}

% \kike{While 360-MLC allows to self-train layout models without annotations, its formulation does not consider any geometry reasoning and treats all layout estimates from every view equally, leading to noisy pseudo labels, particularly for occluded regions and complex scenes.} 
% (see  \cref{fig_banner}-(c)).

\kike{In this paper, we present a self-training framework for room layout estimation that leverages a pre-trained model to compute geometry-aware pseudo-labels for unseen environments. 
% without relying on any geometry assumption, such as Manhattan World assumption, or rooms geometries composed only by planar walls.
Our approach utilizes a ray-casting formulation to aggregate multiple noisy estimations along several ray directions for geometry reasoning. Our hypothesis is based on the idea that sampling layout estimates along a ray can locally approximate the probability distribution of the underlying geometry by considering their proximity to the camera view and mutual consistency between views.}
% for occlusion reasoning.}
% Our approach utilizes a ray-casting formulation to aggregate multiple estimations along a ray direction, leveraging their proximity and mutual consistency for occlusion reasoning. This idea rely on the hypothesis that sampling layout estimates along a multiple rays can locally approximate the probability distribution of the underlying geometry by considering multiple views along the scene.}
\kike{This simple yet effective approach yields remarkable room geometry definitions, including shapes with circular and non-planar walls, as well as effectively handling occluded geometries. See~\cref{fig_teaser}-(b).}

% In this paper, we present MLC++, a geometry-aware self-training framework for room layout estimation that leverages a pre-trained model to produce its own pseudo-labels in unlabeled environments.
% (see \cref{fig_banner}-(b)).
% Our hypothesis stems from the idea that sampling layout estimates along a ray can locally approximate the probability distribution of the underlying geometry. Therefore, by sampling it, we can obtain a reliable reference.
% This simple yet effective approach yields remarkable room geometry definitions, including room shapes with circular and non-planar walls. 

To further exploit our proposed solution, we present a Weighted Distance Loss formulation that prioritizes the farthest geometry in the scene during self-training. 
\kike{This stems from the intuition that estimating distant geometries is typically challenging from a single view, suggesting that a multi-view setting may help overcome this issue by considering several complementary views along the scene.}
% This is from the intuition that the farthest geometries are usually challenging to estimate from a single view, suggesting that they offer the most significant information from a multi-view setting.  

\kike{To validate our proposed solution, we collect and label a new dataset (referred to as HM3D-MVL) from HM3D~\cite{hm3d}, particularly addressing occluded, complex, and ample room geometries.
We validate the benefits of the proposed self-training solution through an extensive evaluation in different settings and publicly available datasets~\cite{360_dfpe,zind}, using synthetic and real-world data.}
\kike{Our contributions are as follows:}
\begin{enumerate}

\item \kike{We propose a novel geometry-aware ray-casting formulation for pseudo-labeling unseen scenes directly from the multiple noisy estimations of a pre-trained model.}

\item \kike{We propose a Weighted Distance Loss that exploits the benefits of a multi-view setting by prioritizing distant geometry during self-training.}

\item \kike{We collect and label a new dataset (HM3D-MVL) from~\cite{hm3d}, particularly addressing occluded, complex, and ample room geometry for more diverse scenarios. The dataset and code will be released with this publication.}

\end{enumerate}

\section{Related Work}
\vspace{-2mm}
\label{sec:related_works}
\vspace{-1mm}
\subsubsection{Room Layout Estimation.}
\kike{Estimating the room layout geometry is a long-standing problem, where earlier works \cite{tsai2011real, chao2013layout, xu2017pano2cad} mainly rely on key features, semantic cues, and prior geometries to reason about the underlying geometry.}
\kike{While deep learning solutions for this task have brought robustness in the estimation by leveraging supervision from labeled data~\cite{lee2017roomnet, layoutnet, fernandez2020corners, dula}, most of these solutions define the problem as a regression map task. An outstanding solution that changes this paradigm is HorizonNet~\cite{horizonnet}, which redefines the optimization as an 1D boundary regression problem, simplifying the definition for the layout geometry. Upon this solution, approaches like~\cite{hohonet} have impressive results by leveraging a simple layout definition. Another advance is LED2Net~\cite{led2net} and LGTNet~\cite{lgtnet}, which introduces a horizon-depth vector definition, constraining the layout geometry directly on Euclidean space. Upon this solution, recent approaches~\cite{DMHNet, U2rle} present further constrains during training, none of them targeting multi-view consistency.}

\vspace{-5mm}
\subsubsection{Multi-view Layout.}
\kike{Recent approaches in multi-view setting~\cite{graph_covis,gprnet, CoVisPose} define the multi-view layout estimation problem jointly with camera pose registration. In particular, \cite{CoVisPose} introduces important concepts for geometry reasoning, such as layout occlusion and layout match correspondences strictly relying on ground truth annotations. An outstanding solution in this manner is Graph-Covis~\cite{graph_covis}, which is built upon~\cite{CoVisPose} to define a multi-view setting capable of estimating layout and camera pose from multi-views using a graph neural network approach. Nevertheless, these solutions rely on ground truth annotations for reasoning the underlying geometry.} 

\vspace{-5mm}
\subsubsection{Semi-Supervised and Self-training Layout Estimation.}
Semi-supervision and self-training methods aim to define a reliable reference to constrain the learning optimization without ground truth annotations~\cite{lee2013pseudo}. 
Along this line, SSLayout360~\cite{sslayout} utilizes a Mean Teacher framework \cite{tarvainen2017mean} to train a layout estimation model using pseudo-labels from a exponential-moving-average operation. However, \cite{sslayout} treats each image in isolation, neglecting valuable geometric information from alternate camera views. Furthermore, the challenge arises from the inherent noise in pseudo labels. Existing approaches aim to mitigate this noise through techniques such as assembling predictions across diverse augmentations \cite{berthelot2019mixmatch, radosavovic2018data} or by selectively retaining only those pseudo-labels with high confidence \cite{sohn2020fixmatch}.

On the other hand, a practical solution for self-training models is to leverage information from a pre-trained model. In 360-MLC~\cite{360_mlc}, multiple estimations of a pre-trained model \cite{horizonnet} are re-projected into a camera view from which pseudo labels are sampled. However, this formulation does not consider any geometry prior and treats every geometry estimation equally, which yields noisy labels, particularly for occluded geometry. To the best of our knowledge, a self-training formulation that handles geometry in a multi-view setting without relying on label annotation has not been studied. 
\vspace{-3mm}
\begin{figure}[t!]
\centering
  \includegraphics[ width=\linewidth]
{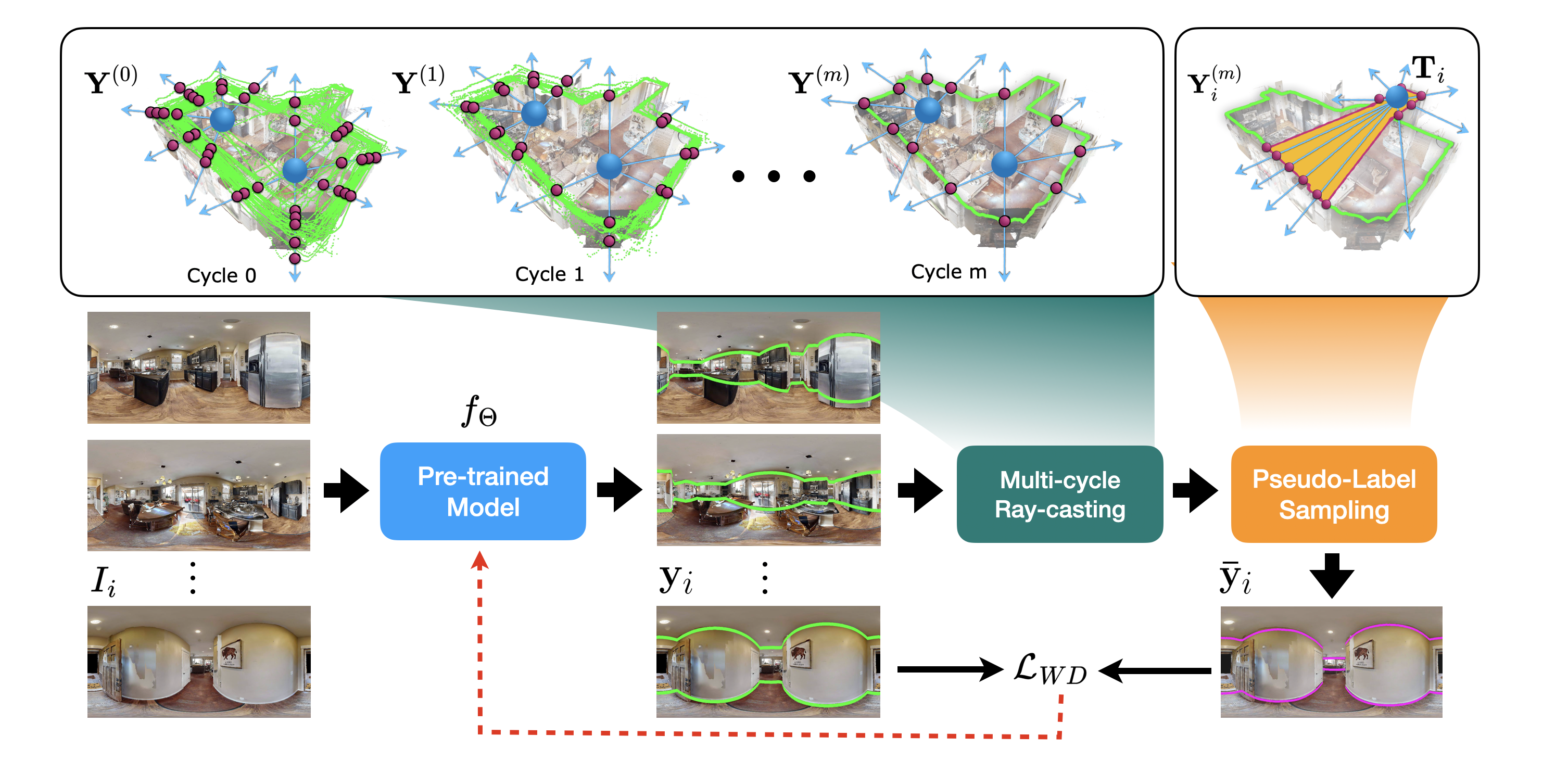}
% \vspace{-2mm}
    \caption{
    \textbf{Self-training Pipeline.}
        \kike{We use a pre-trained model $f_{\Theta}$ to estimate multiple layouts~$\mathbf{y}_i$ from multiple views $I_i$ in an unseen scene. We aggregate all noisy estimates ~$\mathbf{Y}^{(0)}=\mathrm{concat}(\{\mathbf{y}_i\}_{i:n})$ using our proposed Multi-cycle ray-casting process. Then, we sample our pseudo-label $\mathbf{\bar{y}}_i$ at the 
        camera position $\mathbf{T}_i$ from the filtered set of layouts $\mathbf{Y}_i^{(m)}$. Finally, we constraint our self-training optimization using our proposed Weighted-distance loss $\mathcal{L}_{WD}$.} 
        % \textbf{MLC++ for self-training room layout models.} By leveraging multiple estimations from a pre-trained model, as depicted in (a), MLC++ is capable of defining remarkable pseudo-labels through a ray-casting data aggregation process from every view in the scene, as illustrated in (b). 
        % The result of this process is a filtered room geometry utilized to sample pseudo labels for each camera view in the scene, illustrated in (c) as a green sphere. Subsequently, these pseudo-labels are employed for self-training the pre-trained model (see \cref{sec:method}).
        } 
    \label{fig_pipeline}
    \vspace{-5mm}
\end{figure}

\section{Proposed Method}\label{sec:method}
\vspace{-2mm}
\kike{The following outlines our proposed self-training framework for room layout estimation. 
\kike{In~\cref{sec:self-training}, we describe the multi-view layout consistency problem (MLC) as well as the preliminaries for self-training room layout models.}
% In section \cref{sec:self-training}, we describe the room layout estimation setting and define the self-training optimization through pseudo-labeling. 
In~\cref{sec:ray_casting}, we present our ray-casting data aggregation process to create geometry-aware pseudo-labels solely from estimated data. 
Lastly, in~\cref{sec:w_fucntion}, we present our weighted loss formulation towards leveraging the farthest distant geometry in a scene. For illustration purposes, an overview of our self-training framework is depicted in \cref{fig_pipeline}.}
\vspace{-3mm}
\subsection{Self-training Room Layout with Multi-view Layout Consistency}\label{sec:self-training}

\kike{In general, self-training a room layout model by multi-view layout consistency (MLC) aims to fine-tune a pre-trained model with reliable pseudo-labels computed from multiple estimations along an unseen scene~\cite{360_mlc}. This scene with $n$ views can be defined as follows:}
\begin{equation} \label{eq_scene}
\mathcal{S} = \{(I_i, \mathbf{T}_i)\}_{i=1:n}~,~~I_i \in \mathbb{R}^{H \times W}~,~\mathbf{T}_i\in SE(3)~,
\end{equation} 
\kike{where $\mathcal{S}$ is the set of inputs views,~$I_i$ represents a panoramic image of size $H \times W$ pixels, and $\mathbf{T}_i$ is the corresponding camera pose with rotation $\mathbf{R}_i\in SO(3)$ and translation $\mathbf{t}_i\in \mathbb{R}^3$ defined in world coordinates. For any view in the set $\mathcal{S}$, we can define an estimated layout geometry as follows:}
\begin{gather}
  \begin{aligned}
    \mathbf{y}_i&= \pi(f_\Theta(I_i), \mathbf{T}_i), & \mathbf{y}_i \in \mathbb{R}^{3 \times W},
  \end{aligned}\label{eq_ly_est}
\end{gather}
\kike{where $f_\Theta$ is a layout model parameterized by $\Theta$, $\pi(\cdot)$ is a projection function that transforms the model's prediction into the Euclidean space, and $\mathbf{y}_i$ is the estimated layout geometry registered in world coordinates. For simplicity, we refer to $\mathbf{y}_i$ as the floor boundary only.  For 
layout models such as \cite{hohonet, horizonnet}, $\pi(\cdot)$ processes a 1D boundary vector defined in spherical coordinates, while models~\cite{led2net, lgtnet} handle a 1D horizon-depth estimation. A closed-form definition for both is described in our supplementary material.}

\kike{By estimating multiple layouts from every view in the scene, we can define the pseudo labeling process as follows:}
% \begin{equation} \label{eq_ly_concat}
% % \mathbf{Y} =\mathbf{y}_1 \cup \cdots \cup \mathbf{y}_n~,~~ \mathbf{Y} \in \mathbb{R}^{3 \times nW}~,
% \mathbf{Y} =\mathrm{concat}(\{\mathbf{y}_0, \cdots, \mathbf{y}_n\})~,~~ \mathbf{Y} \in \mathbb{R}^{3 \times nW}~,
% \end{equation}
% \begin{equation}
% \label{eq_Y_in_cc}
% \mathbf{Y}_i = \mathbf{R}_i \mathbf{Y} + \mathbf{t}_i,~~~
% \bar{\mathbf{y}}_i = \Phi(\mathbf{Y}_i),~~~\bar{\mathbf{y}}_i \in \mathbb{R}^{3 \times W},
% \end{equation}
\begin{gather}
  \begin{aligned}
    & \mathbf{Y} =\mathrm{concat}(\{\mathbf{y}_0, \cdots, \mathbf{y}_n\})~,~~ \mathbf{Y} \in \mathbb{R}^{3 \times nW}~,\\
    & 
    \mathbf{Y}_i = \mathbf{R}_i \mathbf{Y} + \mathbf{t}_i,~~~
\bar{\mathbf{y}}_i = \Phi(\mathbf{Y}_i),~~~\bar{\mathbf{y}}_i \in \mathbb{R}^{3 \times W},
  \end{aligned}\label{eq_Y_in_cc}
\end{gather}

\kike{where $\mathbf{Y}$ 
% \in \mathbb{R}^{3\times nW}$ 
is the concatenation of $n$ layout geometries estimated by \cref{eq_ly_est}, $\mathbf{Y}_i$ stands as the rigid transformation of $\mathbf{Y}$ into the $i-$th camera reference, and $\Phi(\cdot)$ is the aggregating function that estimates a pseudo-label $\bar{\mathbf{y}}_i$ for the $i-$th view in the scene.}

\kike{Note that, in the case of 360-MLC~\cite{360_mlc}, $\Phi(\cdot)$ is the function that samples the median values of re-projected points in the image domain without any geometry reasoning, see \cref{fig_teaser}-(c). In \cref{sec:ray_casting}, we redefine $\Phi(\cdot)$ as a ray-casting function for computing geometry-aware pseudo-labels.}
% as illustrated in \cref{fig_banner}-(c).}

\kike{The self-training optimization of $f_\Theta$ with multiple pseudo-labels $\bar{\mathbf{y}}_i$ can be defined as follows:}
\begin{equation} \label{eq_self_training_opt}
\mathop{\text{min}}\limits_{\Theta}  \frac{1}{n}\sum_{i=1}^{n}{\omega_i \cdot \mathcal{L} \left(~  f_\Theta(I_i),~\pi^{-1}(\bar{\mathbf{y}}_i\right))}, 
\end{equation}
\kike{where $\pi^{-1}(\cdot)$ is the inverse function presented in \cref{eq_ly_est}, $\omega_i \in \mathbb{R}^{W}$ is a weighted vector associated to the uncertainty in each pseudo-label~$\bar{\mathbf{y}}_i$, and $\mathcal{L}(\cdot)$ is the loss function that constraints the self-training optimization.}

\kike{Note that, in the case of 
360-MLC~\cite{360_mlc}, The self-training constraint is defined as a weighted L1 loss with $\omega_i=\sigma_i^{-2}$, where $\sigma_i$ is the standard deviation of re-protected points in the image domain.}
\kike{In \cref{sec:w_fucntion}, we redefine $\omega_i$ into our weighted-distance function that prioritizes distance geometries from the camera view during self-training.}
\vspace{-3mm}

\subsection{Pseudo-labeling by Ray-casting}\label{sec:ray_casting}
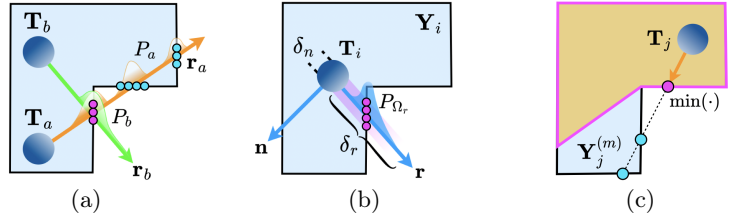
\begin{figure}[t!]
  \centering
\includegraphics[width=10cm]
{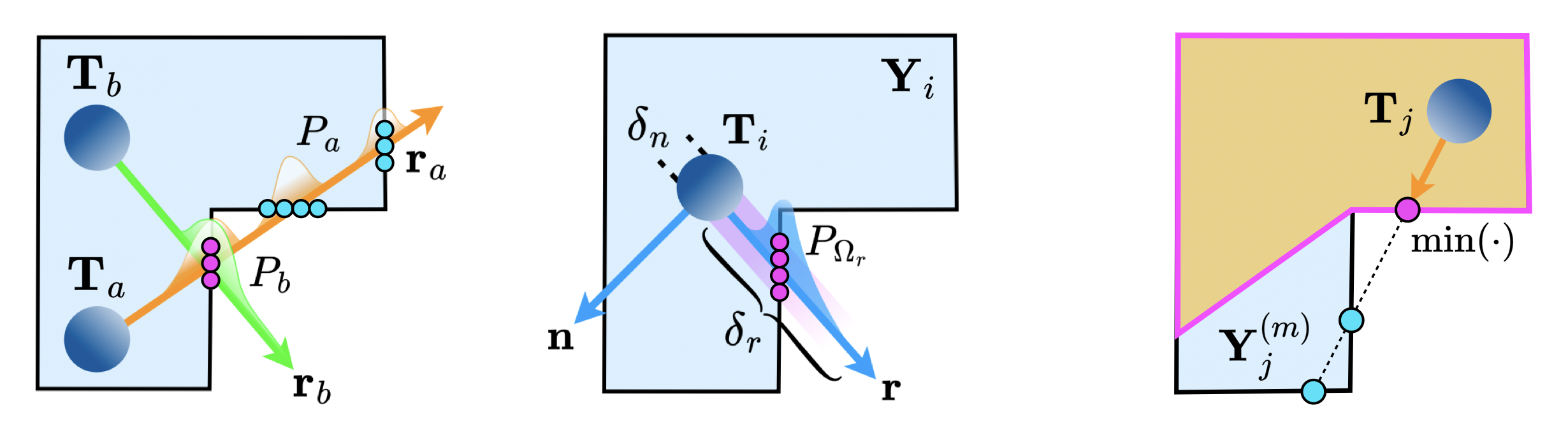}
    \begin{picture}(0, 0)
    \put(-258,-3.8){(a)}
    % from different camera views}
    \put(-153,-3.8){(b)}
    \put(-48,-3.8){(c)}
    \end{picture}
  \caption{\textbf{Ray-Casting:} In panel (a), different ray directions from different camera views are shown. Note that due to occluded geometries and different camera positions, the probability distribution along a ray may vary significantly. In panel (b), one of our constraints to handle occluded geometries is depicted, i.e., sampling a nearby region along the ray to define $P_{\Omega_r}$. In Panel (c), we sample a pseudo-label (magnet contour) from a filtered layout boundary $\mathbf{Y}^{(m)}_j$ at the camera $\mathbf{T}_j$ by using $\mathrm{min}(\cdot)$ function to sample the non-occluded points on the rays (see \cref{sec:ray_casting}).}
  \label{fig_ray_casting}
  \vspace{-5mm}
\end{figure}

% \vspace{-2mm}
\subsubsection{Probability distribution on a ray.}
\kike{We hypothesize that the projection of multiple layout estimates onto a ray can describe a probability distribution of the underlying geometry. This distribution can then serve as the basis for sampling reliable pseudo-labels.}
\kike{To this end, we propose a ray-casting formulation that projects multiple estimates of a pre-trained model into a set of ray directions defined in the bird-eye-view (BEV), i.e., ray vectors defined in the xz Euclidean plane. This is motivated by previous works \cite{led2net, lgtnet} to represent a room layout geometry directly in the Euclidean space, avoiding distortion and discrete issues presented in the image domain.}
% \kike{The main goal of a pseudo-label is to define a reliable reference for self-training a model in a new data domain~\cite{360_mlc,lee2013pseudo,sslayout}. To this end, we propose a ray-casting formulation that projects and aggregates multiple layout estimates of a pre-trained model into a set of ray directions
% \kike{defined in the bird-eye-view (BEV), i.e., ray vectors defined in the xz Euclidean plane. This is motivated from previous works \cite{led2net, lgtnet} to represent a room layout geometry directly in the Euclidean space, avoiding distortion and discrete issues presented in the image domain.}
% \kike{We hypothesize that the projection of layout estimates onto multiple rays can describe a probability distribution of the underlying geometry. This distribution can then serve as the basis for sampling reliable pseudo-labels. To achieve this, we define a set of ray directions in world coordinates as follows:}

\kike{We define a set of ray directions in world coordinates as follows:}
\begin{gather}
  \begin{aligned}
    & \mathcal{R} = \{\mathbf{r}_j\}_{j=1:W }~,~~\mathbf{r}_j \in \mathbb{R}^{3}~,~~|\mathbf{r}_j|=1, \\
    & 
    \mathcal{V} = \{\mathbf{n}_j\}_{j=1:W}~,~~\mathbf{n}_j \in \mathbb{R}^{3}~,~~\mathbf{n}_j\cdot \mathbf{r}^\top_j =\mathbf{0},
  \end{aligned}\label{eq_ray_directions}
\end{gather}

\kike{where $\mathbf{r}_j$ is a ray direction constrained by~$\mathbf{r}_j\cdot[0, 1, 0]^\top=\mathbf{0}$ (i.e., on the xz Euclidean plane), and $\mathbf{n}_j$ is its corresponding normal vector. Then, a pseudo-label from a probability function defined on a ray vector can be defined as follows:}
\begin{equation}
    \label{eq_pr_definition}
    \mathbf{\bar{y}}_{i,r}=\mathbb{E}[P_{r}(\mathbf{Y}_i)]\mathbf{r}, ~~\mathbf{r}\in\mathcal{R},
    % |~\forall\mathbf{r}_j\in\mathcal{R}~\}
\end{equation}
\kike{
where $\mathbf{r}$ is a ray vector introduced by \cref{eq_ray_directions}, $\mathbf{Y}_i$ is the concatenation of all estimated layouts in the $i-$th camera reference as presented in \cref{eq_Y_in_cc}, $\mathbf{\bar{y}}_{i,r}$ stands for the $i-$th pseudo label defined on the ray $\mathbf{r}$, and $P_{r}(\cdot)$ is the unknown probability function along a ray direction $\mathbf{r}$. For simplicity, we refer to this probability function as~$P_r$.
}

\kike{Regardless of the noise within the estimated layout geometries, the density function $P_{r}$ may vary significantly for every camera view and ray direction, in particular for occluded geometry. This phenomenon is illustrated in \cref{fig_ray_casting}-(a), where two density functions $P_{a}$ and $P_{b}$ for the same underlying geometry (magenta dots) are presented. Note that $P_{a}$ defines a multi-modal density function due to multiple occluded geometries (cyan dots), which may lead to a different expectation value compared to $P_{b}$.}
\vspace{-3mm}
\subsubsection{Multi-cycle ray-casting for pseudo-labeling.}
\kike{
To tackle occlusions, we condition $P_{r}$, presented by \cref{eq_pr_definition}, in three ways. First, we increase the sample count near each ray direction and camera view based on the intuition that a higher sample count may enhance the representation of non-occluded geometries. Second, similar to 360-MLC~\cite{360_mlc}, we approximate the expectation of projected samples to $\texttt{median}(\cdot)$ for filtering out noisy estimates, i.e., the median value of points on the ray. However, instead of sampling from a unique view (in the image domain), we sample them from multiple camera locations and ray directions in an iterative process named multi-cycle ray-casting (see \cref{fig_pipeline}). This stems from the fact that sampling over $P_{r}$ from multiple camera locations and directions must yield the same underlying room geometry.
Finally, following the noise reduction, we approximate the expectation of $P_{r}$ to the closest sample on the ray. This is based on the understanding that non-occluded geometries must lie at the closest point along the ray direction. This is illustrated in \cref{fig_ray_casting}-(c), where the pseudo-label for the camera view $\mathbf{T}_j$ (magenta contour) is computed by sampling points on the rays by using the $\texttt{min}(\cdot)$ function.}

\kike{With a slight notation abuse, the projection of nearby estimates onto a ray direction can be defined as follows:}
\begin{gather}
  \begin{aligned}
    &\Omega_{r}(\mathbf{Y}_i)=\{\mathbf{r}\cdot\mathbf{x}^\top ~|~~\forall~\mathbf{x}\in\mathbf{Y}_i\}\quad st.~
    % ,~~\mathbf{r}_j \in \mathcal{R}~,\}~~st.\\ 
    \\&0 < \mathbf{r}\cdot \mathbf{x}^\top \leq \delta_r,
    ~~~\text{and}\quad 
    |\mathbf{n}\cdot \mathbf{x}^\top| \leq \delta_n,
\end{aligned}\label{eq_ray_cast_vecinity}
\end{gather}
\kike{where $\mathbf{x}$ is a 3D-point $\in\mathbb{R}^3$ defined in $\mathbf{Y}_i$, $\mathbf{r}$ and $\mathbf{n}$ are ray-vectors define by \cref{eq_ray_directions}, and $\{\delta_r, \delta_n\}$ is a set of hyper-parameters that allows us to filter out non-local points. This projection is illustrated in \cref{fig_ray_casting}-(b), where the subset of points $\Omega_{r}$ (magenta dots) is defined along the ray vector $\mathbf{r}$. For simplicity, we refer to the probability of these projected samples as $P_{\Omega_{r}}$.}

\kike{The multi-cycle ray-casting process to filter out noisy estimates can be described as follows: }
\begin{gather}
  \begin{aligned}
    &\mathbf{Y}^{(k+1)} =\{\texttt{median}(\Omega_{r_j}(\mathbf{Y}^{(k)}_i))\mathbf{r}_j\}_{i=1:n~~j=1:W}~,
    % &\quad\forall\mathbf{r}_j\in\mathcal{R},\forall\mathbf{T}_i\in\mathcal{S}
\end{aligned}\label{eq_ray_cast_med}
\end{gather}
\kike{where $\mathbf{Y}^{(k)}_i$ stands for the layout estimates in the $i$-th camera reference at the $k$-th cycle. Note that this filtering process is evaluated from all camera views $i$ and all ray directions $\mathbf{r}_j$.}
% \todo{In figure XXX-(a), we exemplify this process with multiple snapshots for a scene in our dataset.}

\kike{Finally, a pseudo label and its uncertainty from a filtered set of layout estimations can be evaluated as follows:}
\begin{gather}
  \begin{aligned}
    &\mathbf{\bar{y}}_i =\{\texttt{min}(\Omega_{{r}_j}(\mathbf{Y}^{(m)}_i))\texttt{r}_j\}_{j=1:W}, \\
    &
    \sigma_i =\{\text{std}(\Omega_{{r}_j}(\mathbf{Y}^{(0)}_i))\}_{j=1:W},
\end{aligned}\label{eq_ray_cast_min}
\end{gather}
\kike{where $\mathbf{Y}^{(m)}_i$ stands for the filtered layout estimates after applying \cref{eq_ray_cast_med} in $m-$th cycles, and $\mathbf{Y}^{(0)}_i$ is the layout estimates before noise reduction. This is because $\sigma_{i}$ aims to describe the underlying noise of the initial layout estimates along the ray directions.}
% \todo{In Fig xxx-(b), we illustrate $\mathbf{\bar{y}}_i$ from different camera locations.}
% \kike{Our pseudo-labeling process is depicted in \cref{alg:ray_cast_alg}.}
% \input{section/3_method/algorithm}
\vspace{-3mm}

\vspace{-2mm}
\subsection{Weighted Distance Loss}\label{sec:w_fucntion}
%Formulation
\begin{figure}[t!]
  \centering
\includegraphics[width=\linewidth]
{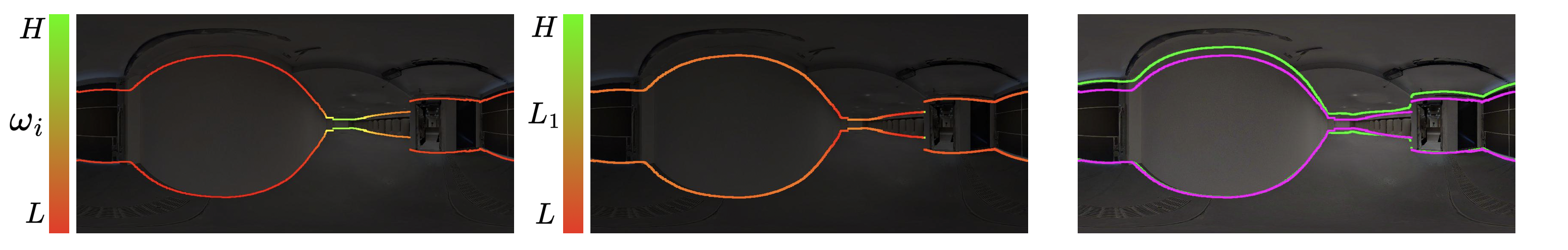}
    \begin{picture}(0, 0)
    \put(-112,0){(a)}
    \put(-2,0){(b)}
    \put(108,0){(c)}
    \end{picture}
  \caption{\textbf{Weighted-distance function:} In panel (a), we illustrate our proposed weighted-distance function $\omega_i$ that prioritizes the farthest geometries in the scene for self-training. In panel (b), under the same scale as (a), we show the $L1$ loss between our proposed pseudo-label and the model estimation. Note that the $L1$ loss evaluation presents a small range w.r.t $\omega_i$ and does not aim at any particular region in the scene. In Panel (c), we present our pseudo-label (magenta line) and the model estimation (green line).}
  \label{fig_loss_wd}
  \vspace{-5mm}
\end{figure}
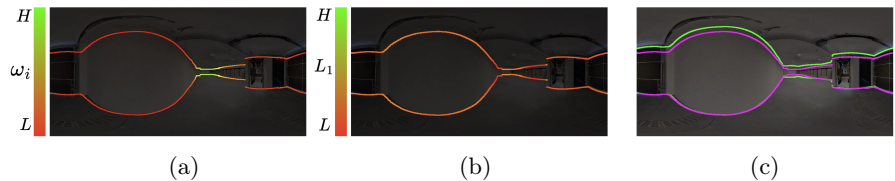
\vspace{-2mm}
\kike{To complement our proposed ray-casting pseudo-labels resented in \cref{sec:ray_casting}, we introduce a weighted loss formulation that particularly focuses on the farthest geometries within a room. This stems from the empirical evidence that pre-trained layout models tend to estimate more accurately the geometries closer to the camera view than those farther away. This limitation can be attributed, in part, to the datasets used for training, e.g., \cite{mp3d, zind}, where room scenes are predominantly captured from the room center, and larger-sized rooms are less represented. Another contributing factor to this limitation is the difficulty in capturing accurate details for the farthest regions from a single view~\cite{lgtnet}. Therefore, we hypothesize that our pseudo-labels may present the most significant impact during self-training when targeting the farthest geometries in a scene.}

\kike{Our weighted formulation can be described as follows:}
% \begin{equation} \label{eq_w_fucntion}
% \omega_i = \frac{e^{\kappa (||\mathbf{\bar{y}}_i|| - d_{min})}}{\sigma^2_i},
% \end{equation}
\begin{equation} \label{eq_w_fucntion}
\mathcal{L}_{WD} = \omega_i||\mathbf{y}_i -\mathbf{\bar{y}}_i||_{1}~~~
\omega_i =\frac{e^{\kappa (||\mathbf{\bar{y}}_i|| - d_{min})}}{\sigma^2_i}
\end{equation}
\kike{where $||\mathbf{\bar{y}}_i||$ is the Euclidean norm of the pseudo labels computed by \cref{eq_ray_cast_min}, $d_{min}$ is the distance from which we want to prioritize the self-training, $\kappa$ is a hyper-parameter that allows us to control the weighting priority to the farthest geometries, and $\sigma_i$ represent the standard deviation computed in \cref{eq_ray_cast_min}.}
\kike{In~\cref{fig_loss_wd}, we compare our proposed weighted-distance function with traditional $L_1$ loss \cite{horizonnet, hohonet, led2net, 360_mlc}. Note that a $L_1$ evaluation does not aim at any particular geometry in the scene, while our proposed $\omega_i$ aims at the farthest walls from the camera view.}
% For illustration purposes, we present in~\cref{fig_loss_wd}-(a) our proposed weighted-distance function $\omega_i$. In~\cref{fig_loss_wd}-(b), a $L_1$ loss evaluation, and\cref{fig_loss_wd}-(c) both our pseudo-label and model estimation boundaries. Note that our proposes weighted distance address the fa  }

\vspace{-5mm}
\section{Experiments}
\label{sec:exp}
\vspace{-2mm}
\subsection{Experimental Setup}
%We first introduce the experiment setup.
\subsubsection{Baseline and Model Backbones.}
\kike{The baseline used in the following experiments is the recent 360-MLC~\cite{360_mlc} taken from the official implementation provided by the authors. For a fair comparison with 360-MLC, we use the same layout model backbone by default, i.e., HorizonNet~\cite{horizonnet} pre-trained in \cite{mp3d}. To further compare our proposed solution, we present results using LGTNet~\cite{lgtnet} pre-trained on~\cite{mp3d} as an additional layout model backbone.}
%We avoid comparison with previous solutions, such as 360-SSLayout~\cite{sslayout}, as our focus is on self-training multi-view layout models without ground truth annotations. 

% Due to the lack of an official implementation of 360-MLC\cite{360_mlc} that handles the LGTNet layout model, we modified the original 360-MLC implementation only for the case of LGTNet. This modification will also be released with our official implementation.}
% To handle different layout geometry definitions in each backbone, we implement a projection for each model.
% (see \cref{sec:self-training} - \cref{eq_ly_est}). 
% For HorizonNet\cite{horizonnet}, this function projects pseudo-labels into spherical coordinates to define a 1D boundary vector. For LGTNet\cite{lgtnet}, it projects pseudo-labels in a 1D horizon-depth vector. A closed-form definition for this projection is described in our supplemental material.}
\vspace{-5mm}
\subsubsection{Datasets.}
\kike{Similar to 360-MLC~\cite{360_mlc}, we show evaluations in the MP3D-FPE dataset~\cite{360_dfpe}. We also show results on the real-world ZInD dataset~\cite{zind}. In addition, we show results in our newly collected dataset rendered from Habitat-v2~\cite{hm3d}, referred to as HM3D-MVL. In the case of the ZInD dataset, we use the layout category ``\textit{visible layout}'' provided by the authors and select the scenes that contain at least five frames per room. 
% This is because our scope focuses specifically on multi-view layout estimation only. 
For all the mentioned datasets, we compute pseudo labels from the training splits, self-train the pre-trained model, and evaluate results on the testing splits using ground truth annotations provided by the authors. To further corroborate our claim of handling occluded geometries, we also present evaluations on a manually selected subset of the testing split that contains samples with geometry occlusions only. We refer to this subset as \textit{Occlusion subset}. Details of these datasets are present in \cref{tab_datasets}.}
\begin{table*}[!t]
% \vspace{-5mm}
\centering
\fontsize{7.0}{9}\selectfont
  \caption{Datasets used in this paper with their statistics, i.e., total frames and average number of frames per room.}
  \vspace{-2mm}
  \label{tab_datasets}
  \begin{tabular}{l c c c | c}
  \toprule
  \multirow{2}{*}{Dataset}
   & Training & Testing & Occlusion & Avg. frames\\
   & set & set & Subset & per room\\
   \midrule
   HM3D-MVL &24344 &2491 &119 &56 \\
   MP3D-FPE~\cite{360_dfpe} &20126 &5254 &157 &46 \\
   ZInD~\cite{zind} &9514 &1157 &191 &6 \\
   \bottomrule
   \end{tabular}
\vspace{-2mm}
\end{table*}
\begin{table*}[!t]
\centering
\fontsize{8.0}{12}\selectfont
  \caption{Quantitative results using the HorizonNet~\cite{horizonnet} backbone. The symbol $\ddagger$ represents that the model is trained with the available labels in the training set, which represents the upper-bound performance.}
  % \vspace{-2mm}
  \label{tab_horizonnet}
  \centering
  \begin{tabular}{l cc | c c || cc | cc }
    \toprule
    & \multicolumn{4}{c}{Testing set} & 
    \multicolumn{4}{c}{Occlusion Subset}\\
    & \multicolumn{2}{c}{2D IoU (\%) $\uparrow$} & \multicolumn{2}{c}{3D IoU (\%) $\uparrow$} & \multicolumn{2}{c}{2D IoU (\%) $\uparrow$} & \multicolumn{2}{c}{3D IoU (\%) $\uparrow$} \\
    
    Method &
    10\% & 100\% &
    10\% & 100\% &
    10\% & 100\% &
    10\% & 100\% \\

    \midrule
    & \multicolumn{8}{c}{Our HM3D-MVL dataset}\\
    \midrule
        Pre-trained~\cite{horizonnet}
        &\multicolumn{2}{c}{76.71} &\multicolumn{2}{c}{71.79} 
        &\multicolumn{2}{c}{78.74}
        &\multicolumn{2}{c}{75.72}
        \\
        360-MLC~\cite{360_mlc}
        &81.69	&82.71	&77.67	&78.71
        &81.66	&79.19	&80.08	&77.72
        \\
        Ours 
        &\textbf{81.74}	&\textbf{82.99}	&\textbf{77.99}	&\textbf{78.95}
        &\textbf{82.05}	&\textbf{83.01}	&\textbf{80.45}	&\textbf{81.38}
        \\

    \midrule
    & \multicolumn{8}{c}{MP3D-FPE dataset~\cite{360_dfpe}}\\
    \midrule
        Pre-trained
        &\multicolumn{2}{c}{77.33}
        &\multicolumn{2}{c}{74.07}
        &\multicolumn{2}{c}{75.09} 
        &\multicolumn{2}{c}{73.36}
        \\
        360-MLC
        &80.84	&80.93	&77.71	&77.69
        &84.15	&84.27	&82.27	&82.04
        \\
        Ours
        &\textbf{81.25}	&\textbf{81.65}	&\textbf{78.15}	&\textbf{78.21}
        &\textbf{85.21}	&\textbf{85.71}	&\textbf{83.16}	&\textbf{83.58}
        \\

    \midrule
    & \multicolumn{8}{c}{ZInD dataset~\cite{zind}}\\
    \midrule
        Pre-trained
        &\multicolumn{2}{c}{68.63}
        &\multicolumn{2}{c}{65.54}
        &\multicolumn{2}{c}{59.98}
        &\multicolumn{2}{c}{53.95}
        \\
        360-MLC
        &74.09	&75.44	&71.21	&72.28
        &62.04	&63.33	&59.29	&60.47
        \\
        Ours
        &\textbf{74.51}	&\textbf{75.71}	&\textbf{72.01}	&\textbf{73.04}
        &\textbf{62.72}	&\textbf{64.01}	&\textbf{60.12}	&\textbf{61.37}
        \\
        \midrule
        \midrule
        Supervised$^\ddagger$ \cite{horizonnet}
        &\multicolumn{2}{c}{84.87}
        &\multicolumn{2}{c}{81.55}
        &\multicolumn{2}{c}{79.44}
        &\multicolumn{2}{c}{75.56}
        \\
    \bottomrule
    % \multicolumn{9}{c}{* Supervised~\cite{horizonnet} is the upper-bound performance trained on ZInD~\cite{zind}.}\\
    % \multicolumn{9}{c}{*Pre-trained~\cite{horizonnet} is pre-trained on MP3D~\cite{mp3d}.}
  \end{tabular}
  \vspace{-5mm}
\end{table*}

\vspace{-5mm}
\subsubsection{Evaluation Metrics.}
\kike{
Following~\cite{zou20193d, 360_mlc, lgtnet, horizonnet}, we evaluate results using standard metrics defined for room layout estimation. For room boundary prediction, we evaluate the 2D and 3D intersection-over-union (IoU).
% reporting percentage values. 
For evaluating the smoothness and consistency of layout depth maps, we evaluate root-mean-square (RMS) and $\delta_1$ errors as defined in \cite{horizonnet, lgtnet, led2net}.
All experiments show the median results of 10 self-training runs, each consisting of 15 training epochs.
}
\vspace{-5mm}
\subsubsection{Implementation Details.}
\kike{The layout models' backbones and their pre-trained weights used in our experiments are taken from their official implementation provided by the authors~\cite{horizonnet, lgtnet}. To train the models, we use common data augmentation techniques for the room layout task, i.e., left-right flipping, panoramic rotation, and luminance augmentation. We use the Adam optimizer with a batch size of 4 and a learning rate $1\times10^{-4}$ with a decay ratio of $90\%$. All models are trained on a single Nvidia RTX 2080Ti GPU with 12 GB of memory. For constructing our ray-casting pseudo-labels, we use 15 cycles per room scene, $\delta_r=20$ and $\delta_n=0.01$. For our weighted distance loss function, we use $\kappa=0.5$ and $d_{min}=2$.}
% \vspace{3mm}
\subsection{Quantitative Results}
\subsubsection{Evaluation using HorizonNet Backbone.}\label{sec:hn_experiments}
In these experiments, we compare our proposed ray-casting self-training frameworks with the baseline 360-MLC~\cite{360_mlc}, utilizing the HorizonNet layout model~\cite{horizonnet} pre-trained in~\cite{mp3d}. The results are presented in~\cref{tab_horizonnet} under two main settings: using $10\%$ and $100\%$ of the training set.
Results in the $10\%$ setting show that our proposed solution outperforms 360-MLC, even with a limited number of samples for self-training. Results in the $100\%$ setting further demonstrate the improved performance of our proposed self-training framework.

\kike{By comparing results in the occlusion subset, we find evidence that our solution significantly outperforms 360-MLC. Particularly, while our proposed ray-casting self-training consistently improves performance with increased data, 360-MLC shows only marginal improvement and in some settings, presents a decline in performance. For instance, consider the evaluation of the occlusion subset of the HM3D-MVL dataset. When using only $10\%$ of the data, 360-MLC achieves $81.66\%$ 2D IoU. However, the result on the $100\%$ setting shows a drop in performance to $79.19\%$. This suggests that 360-MLC contains a large amount of noisy pseudo labels such that increasing the amount of data significantly hurts the performance. We argue that the general benefit of our ray-casting pseudo-labels is mainly due to their strong reasoning capability on occluded geometries.}
\kike{Additionally, we present a comparison against the fully-supervised HorizonNet~\cite{horizonnet} on ZInD~\cite{zind} as an upper-bound references. Although our proposed ray-casting framework effectively self-train a pre-trained model into a new domain, we still found a gap when using manual labels, showing potential direction for future works.}

\vspace{-2mm}
\subsubsection{Evaluation using LGTNet Backbone.}
\begin{table*}[!t]
\centering
\fontsize{8.0}{12}\selectfont
  \caption{Quantitative results using the LGTNet~\cite{lgtnet} backbone. The symbol $\ddagger$ represents that the model is trained with the available labels in the training set, which represents the upper-bound performance.}
  \vspace{-2mm}
  \label{tab_lgt_net}
  \centering
  \begin{tabular}{l cc  cc || cc  cc }
    \toprule
    & \multicolumn{4}{c}{Testing set} & 
    \multicolumn{4}{c}{Occlusion Subset}\\
     
    Method &
    2D IoU $\uparrow$ & 3D IoU $\uparrow$ &
    RMS $\downarrow$& $\delta_1$ $\uparrow$ &
    2D IoU $\uparrow$ & 3D IoU $\uparrow$ &
    RMS $\downarrow$& $\delta_1$ $\uparrow$\\

    \midrule
    & \multicolumn{8}{c}{Our HM3D-MVL dataset}\\
    \midrule
    % \multirow{3}{*}{HM3D-MVL}
        pre-trained~\cite{lgtnet}
        &78.90	
        &74.04	
        &0.409	 
        &0.864	
        &80.22	&78.10	&0.2784	 &0.931\\
        360-MLC~\cite{360_mlc}
        &84.07	&78.85	&0.394	&0.897
        &71.29	&68.54	&0.573	&0.884\\
        Ours 
        &\textbf{86.49}	&\textbf{81.90}	&\textbf{0.293}	&\textbf{0.913}
        &\textbf{83.75}	&\textbf{82.06}	&\textbf{0.264}	&\textbf{0.950}\\

    \midrule
    & \multicolumn{8}{c}{MP3D-FPE Dataset~\cite{360_dfpe}}\\
    \midrule
        pre-trained
        &79.66	&76.32	&0.324
        &0.892
        &78.22	&76.39	&0.243
        &0.949\\
        360-MLC
        &82.99	&77.22	&0.358	&0.883
        &79.16	&75.07	&0.378	&0.907\\
        Ours 
        &\textbf{85.69}	&\textbf{81.80}	&\textbf{0.242}	&\textbf{0.931}	
        &\textbf{86.33}	&\textbf{84.27}	&\textbf{0.168}	&\textbf{0.963}\\

    \midrule
    & \multicolumn{8}{c}{ZInD dataset~\cite{zind}}\\
    \midrule
    
    % \multirow{2}{*}{ZInD~\cite{zind}}
        pre-trained
        &72.59	&69.67	&0.445	&0.897
        &60.30	&57.51	&0.645	&0.846\\
        % & 360-MLC
        % &-	&-	&-	&-
        % &-	&-	&-	&-\\
        Ours
        &\textbf{76.77}	&\textbf{74.42}	&\textbf{0.406}	&\textbf{0.905}
        &\textbf{64.76}	&\textbf{62.38}	&\textbf{0.593}	&\textbf{0.857}
        \\
        \midrule
        \midrule
        Supervised$^\ddagger$ \cite{lgtnet} 
        &{87.64}
        &{84.61}
        &{0.286}
        &{0.931}
        &{80.51}
        &{77.87}
        &{0.393}
        &{0.873}
        \\
    \bottomrule
    % \multicolumn{9}{c}{*Supervised~\cite{lgtnet} represents the upper-bound performance trained on ZInD~\cite{zind}.}\\
    % \multicolumn{9}{c}{*The pre-trained~\cite{lgtnet} is pre-trained on MP3D~\cite{mp3d}.}
  \end{tabular}
  \vspace{-5mm}
\end{table*}
\kike{In this experiment, we aim to validate the performance of our proposed solution compared to 360-MLC when utilizing a state-of-the-art solution for room layout estimation, i.e., LGTNet~\cite{lgtnet}. The results are depicted in \cref{tab_lgt_net}. Although a robust backbone model benefits both models, our self-training framework significantly outperforms 360-MLC across all evaluations. Hence corroborating the versatility of our solution by leveraging new room layout formulations. Results of 360-MLC in the ZInD dataset were omitted due to several failures during self-training, we argue that this is due to the limitation of 360-MLC to handle a setting with a few number frames and horizon-depth constrain. Similar to the experiment presented in \cref{tab_horizonnet}, We present upper-bound results that provide evidence of a gap between training on manual annotations and pseudo-labels, indicating a potential direction for future work.}
\vspace{-2mm}
\subsection{Qualitative Results}
\begin{figure}[t!]
\centering
  \includegraphics[width=10cm]
{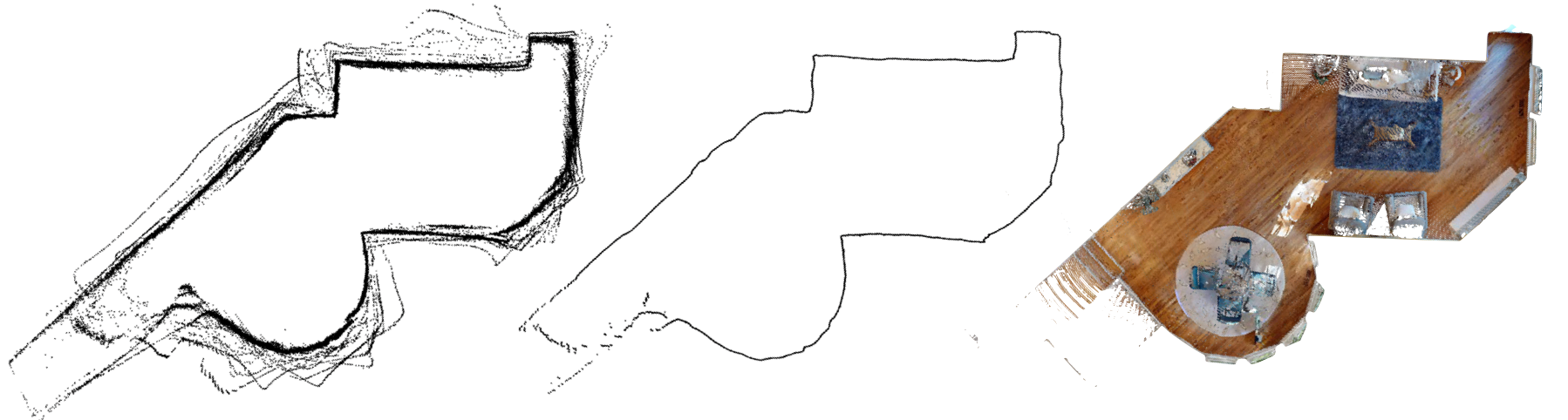}
    \put(-280,-7){(a) 360-MLC~\cite{360_mlc}}
    \put(-165,-7){(b) Ours}
    \put(-100,-7){(c) Point Cloud reference}
    \caption{
        \textbf{Qualitative comparisons of estimated pseudo-labels.} We show a BEV projection of all pseudo-labels for the scene: (a) pseudo-labels from 360-MLC~\cite{360_mlc}, (b) pseudo-labels from our proposed multi-cycle ray-casting, and (c) Point cloud for reference purposes.}
    \label{fig_qualitative_bev}
\vspace{-5mm}
\end{figure}

\subsubsection{Qualitative Results on Panoramic Images.}
\kike{For illustration purposes, we present in~\cref{fig_qualitative_pano} several qualitative results of our proposed self-training framework compared with 360-MLC. Based on these results, we find that our solution shows a significant improvement in handling occluded geometries in all datasets. 
In addition, we observe that our self-training formulation consistently provides more accurate estimations of geometries near entrances and gates. We argue that this is due to the effectiveness of our ray-casting pseudo-labels in defining reliable room geometry, even for those challenging view locations.}
\vspace{-5mm}

\subsubsection{Qualitative Pseudo-labels Results.}
\begin{figure*}
    \small
    \centering
    \footnotesize
    \begin{tabular}{c c c c}
    
        % ############################## HM3D-mvl
        
        \includegraphics[page=9, width=0.22\linewidth]{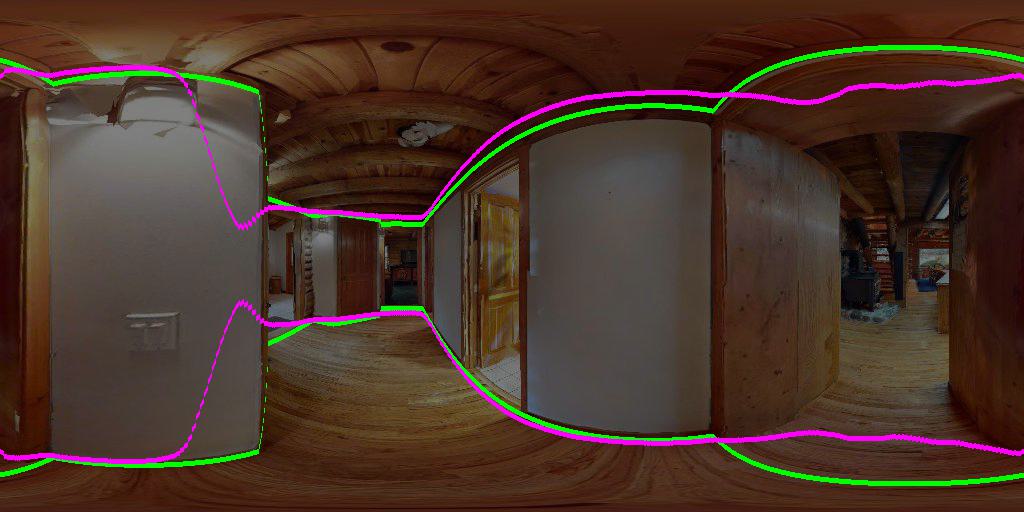}

        \includegraphics[page=9, width=0.22\linewidth]{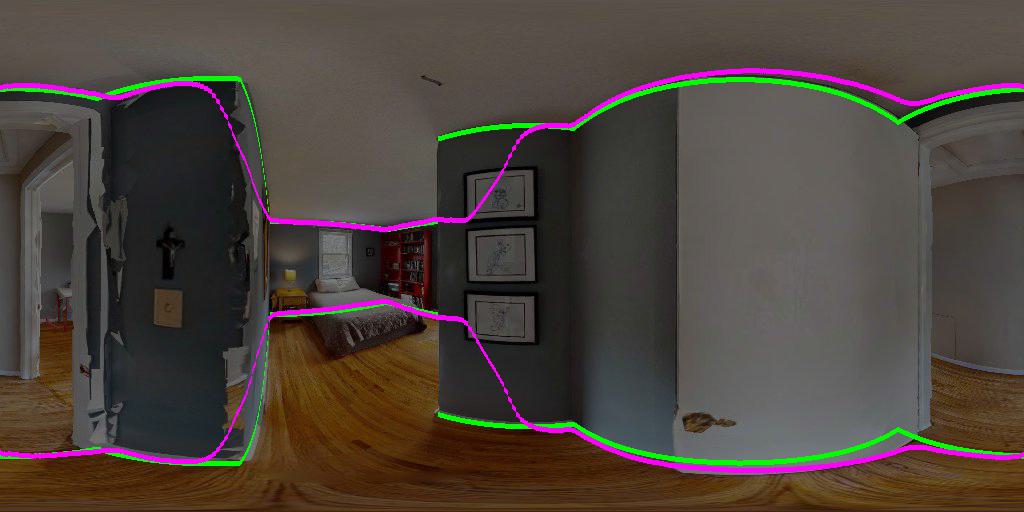}
        
        \includegraphics[page=7, width=0.22\linewidth]{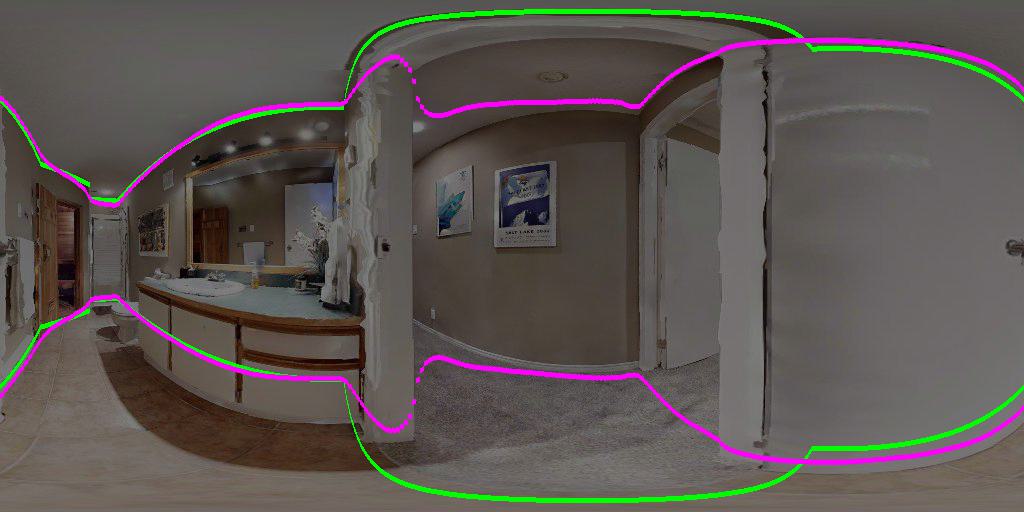}
        
        \includegraphics[page=4, width=0.22\linewidth]{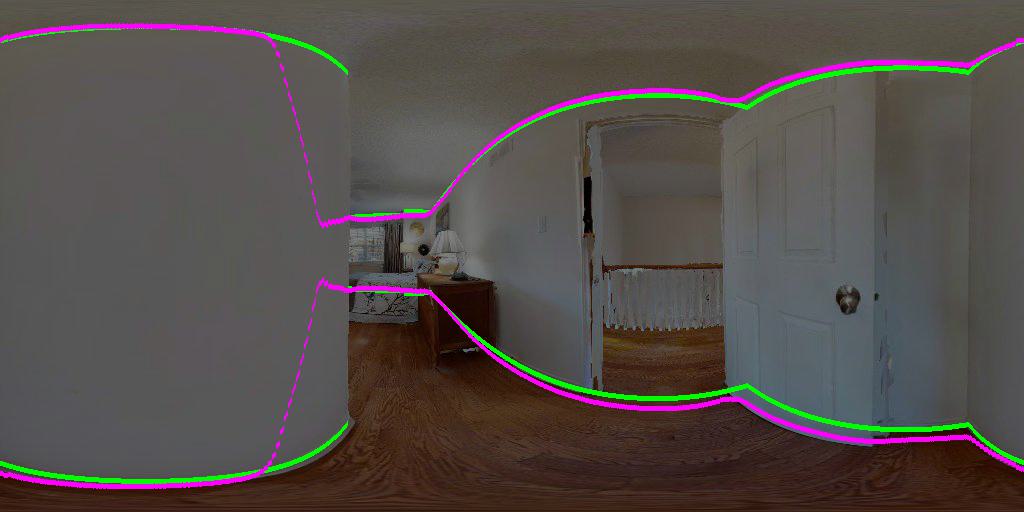}
        
        \\
        \includegraphics[page=9, width=0.22\linewidth]{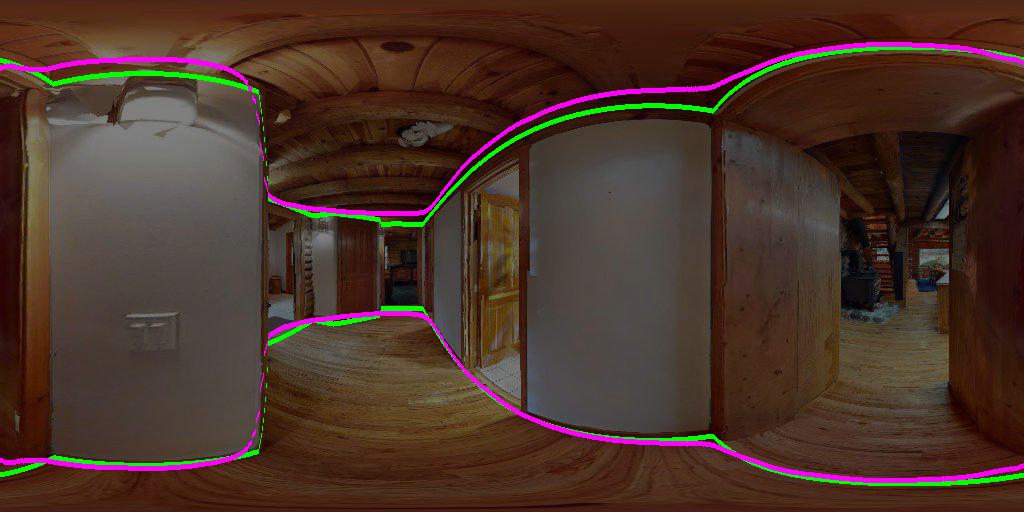}

        \includegraphics[page=9, width=0.22\linewidth]{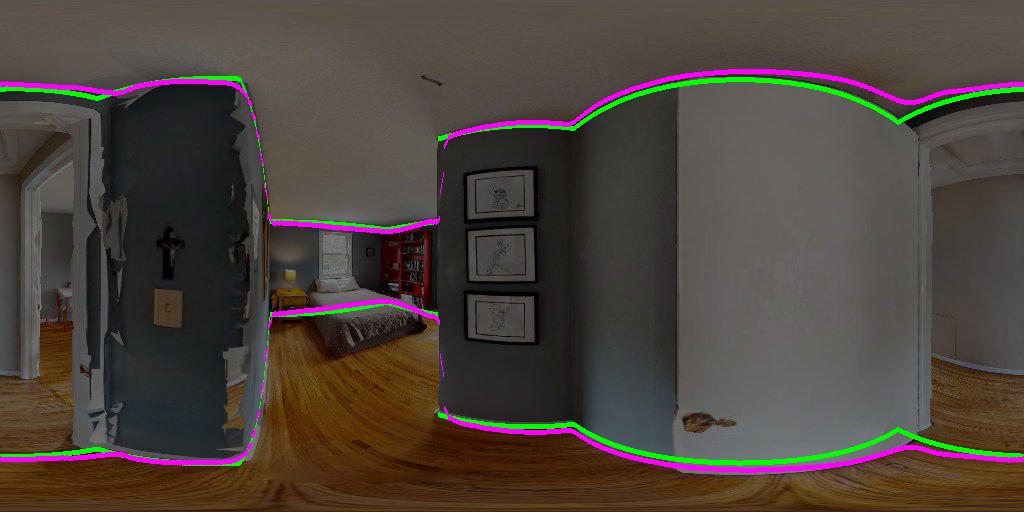}
        
        \includegraphics[page=7, width=0.22\linewidth]{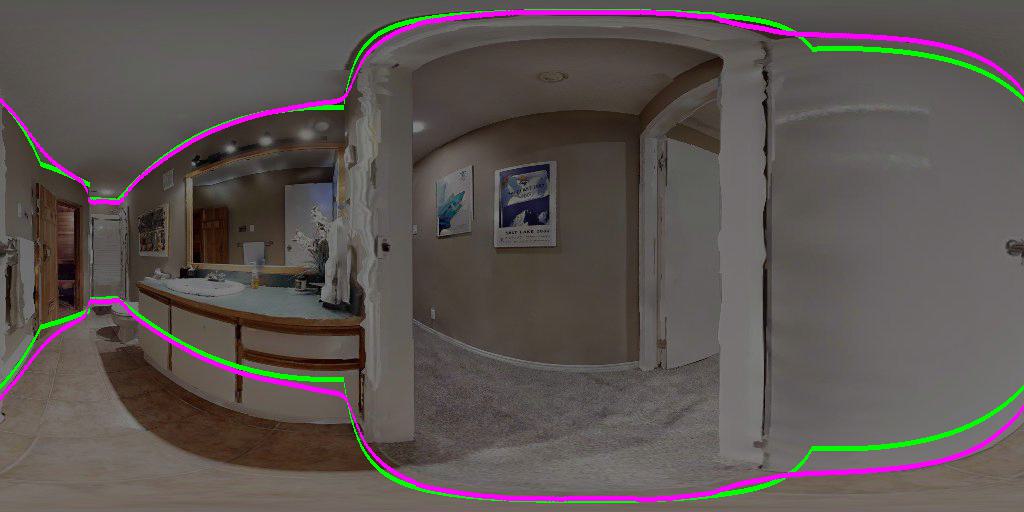}
        
        \includegraphics[page=4, width=0.22\linewidth]{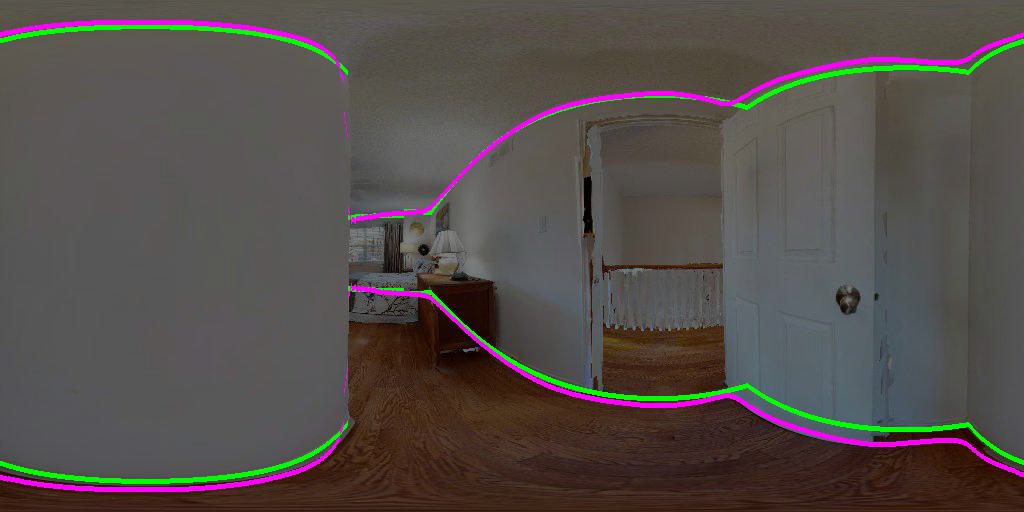}
        \\
        \\
        \\
        % ############################## 
        % MP3D-FPE
        
        \includegraphics[page=10, width=0.22\linewidth]{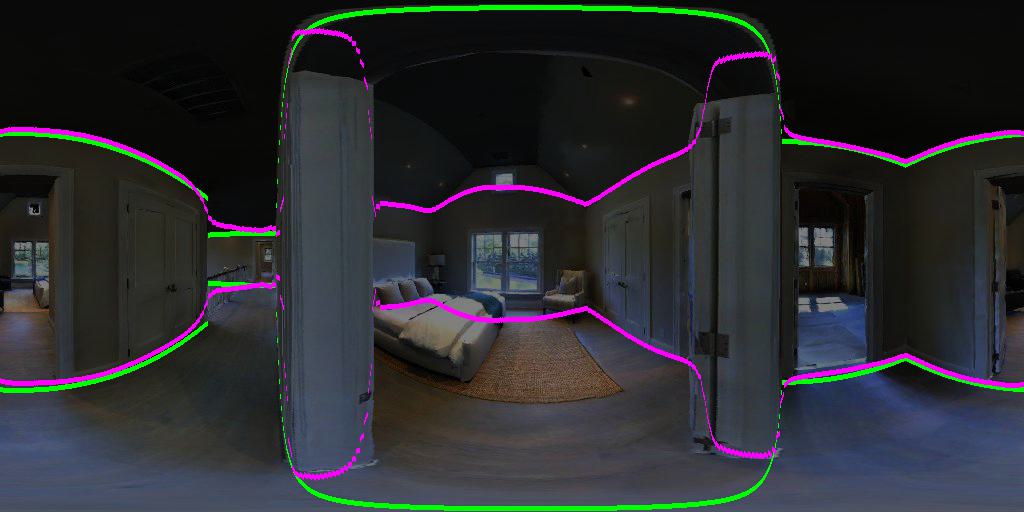}

        \includegraphics[page=8, width=0.22\linewidth]{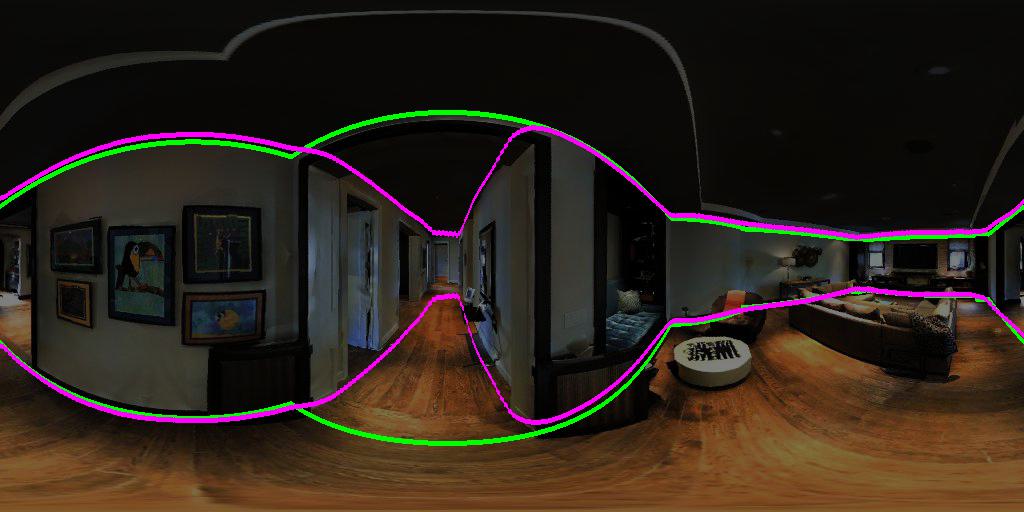}

        \includegraphics[page=8, width=0.22\linewidth]{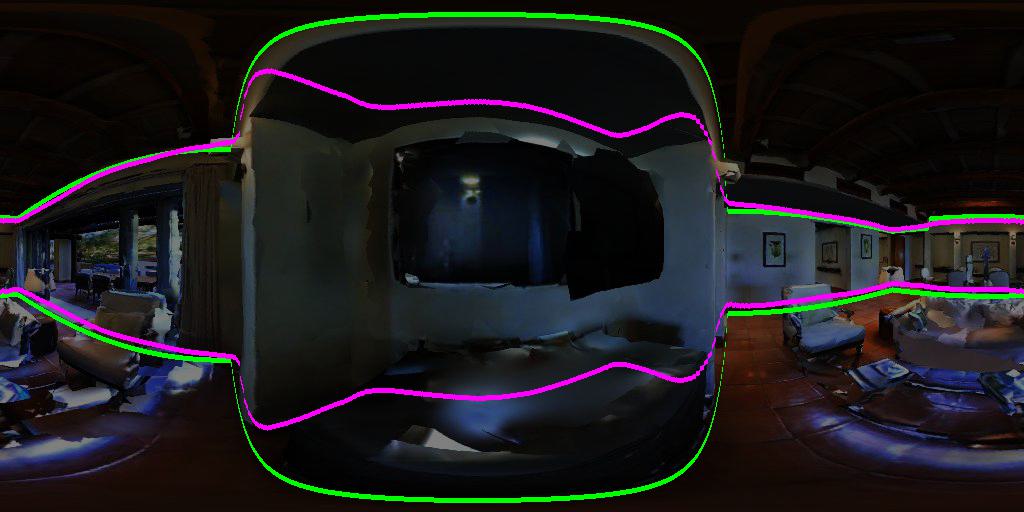}

        \includegraphics[page=2, width=0.22\linewidth]{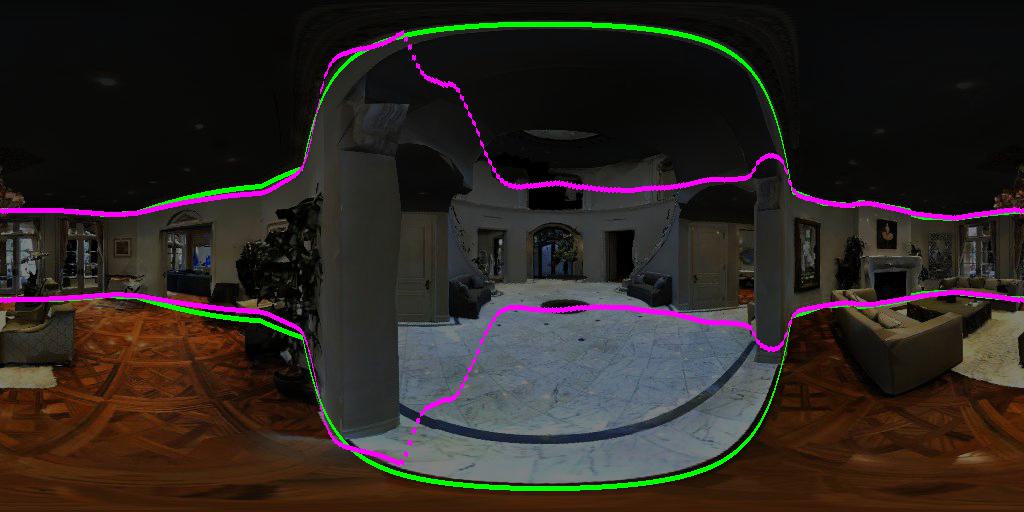} 
        \\
        \includegraphics[page=10, width=0.22\linewidth]{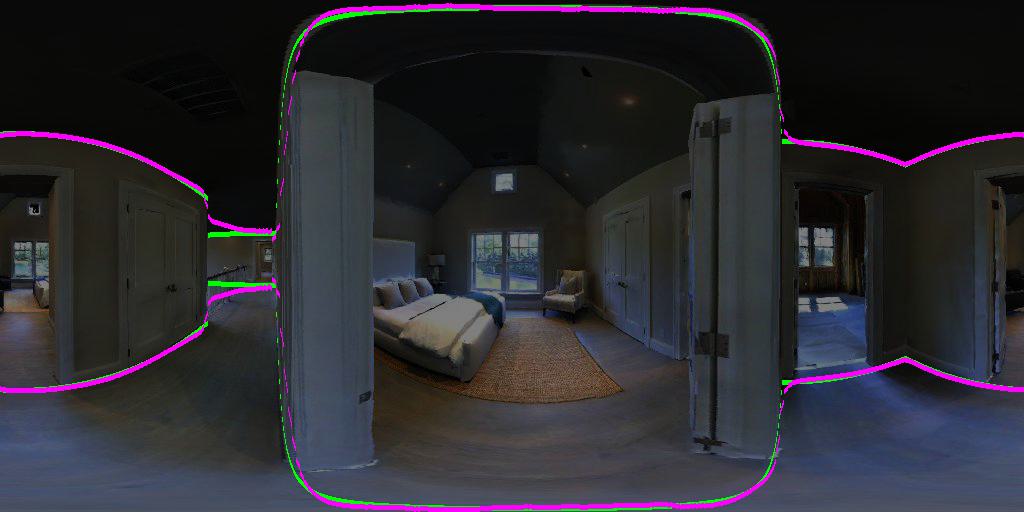}

        \includegraphics[page=8, width=0.22\linewidth]{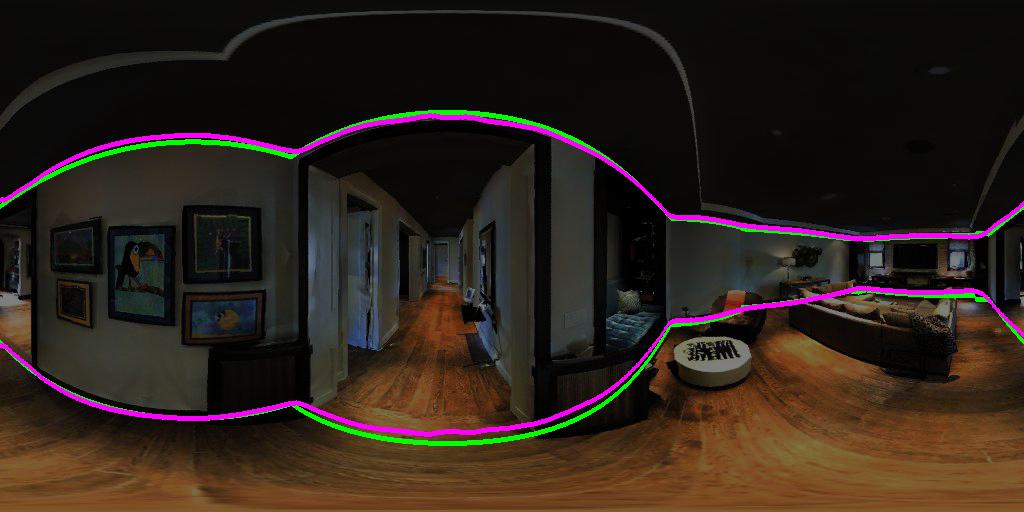}

         \includegraphics[page=8, width=0.22\linewidth]{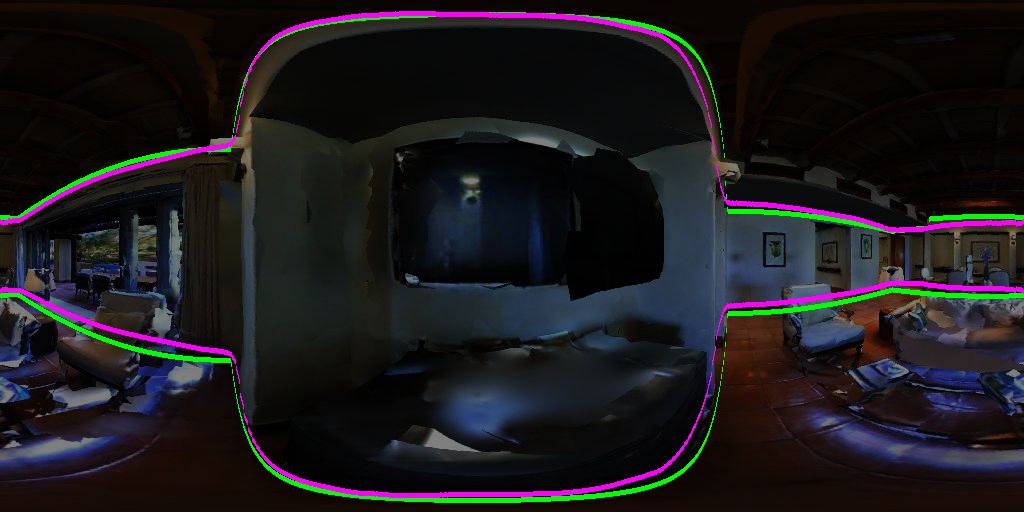}

        \includegraphics[page=2, width=0.22\linewidth]{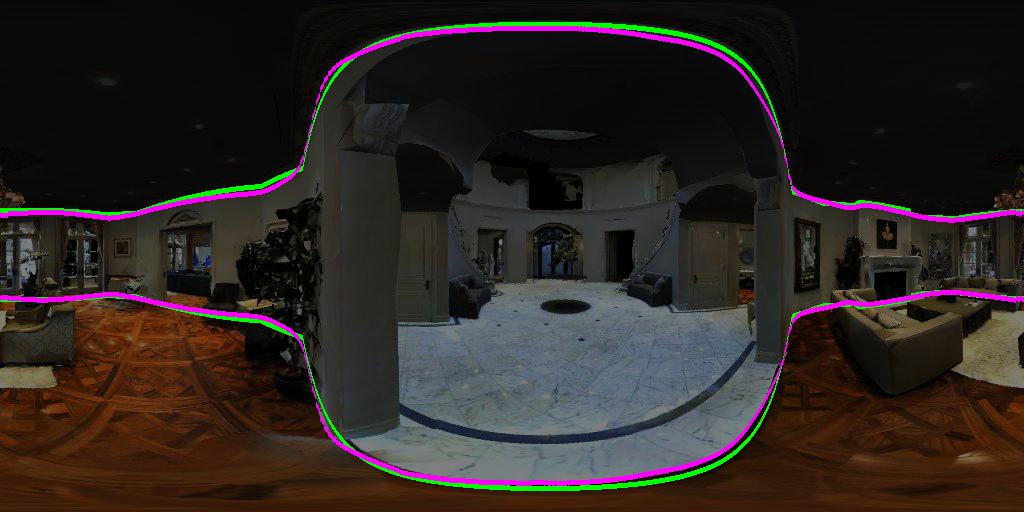} 
        \\
        \\
        \\

        % ###########################
        % ZIND
        
        \includegraphics[page=11, width=0.22\linewidth]{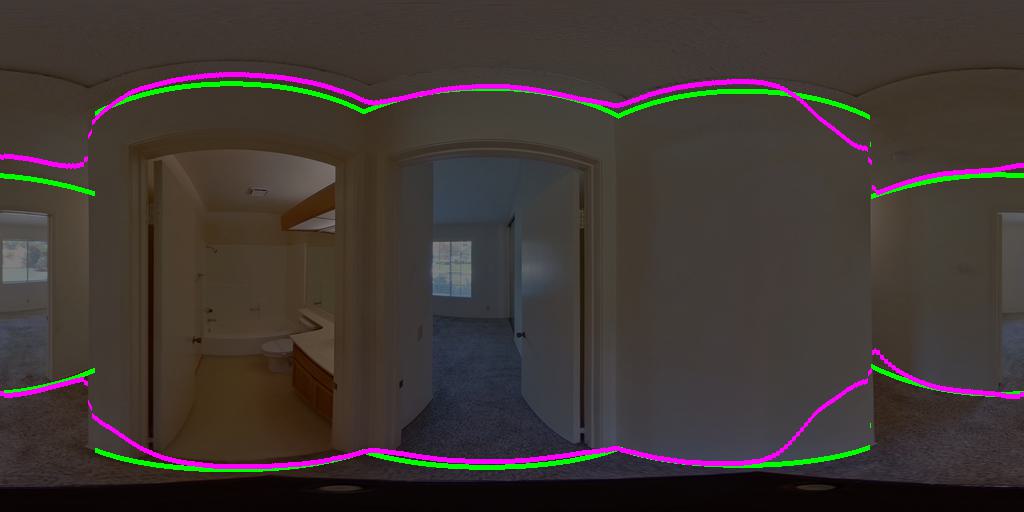}

        \includegraphics[page=13, width=0.22\linewidth]{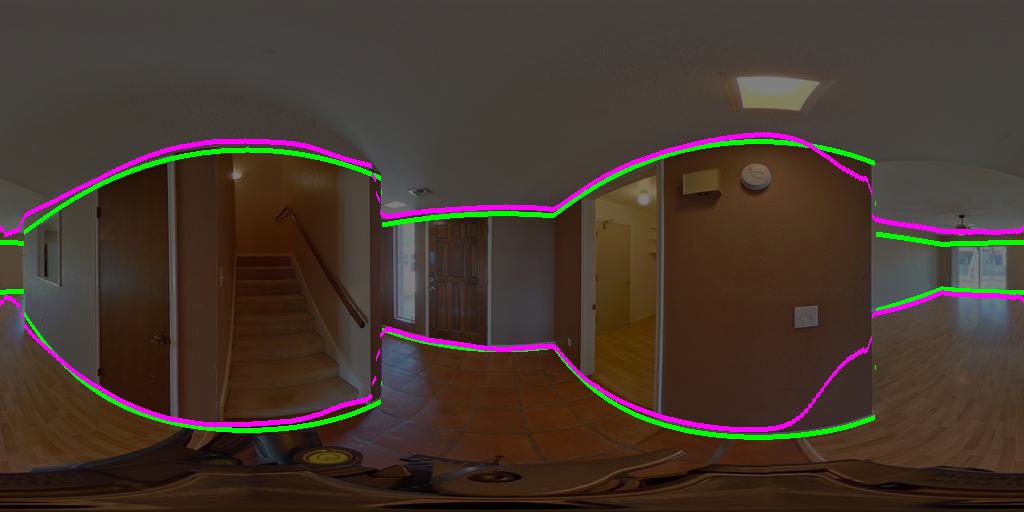}
        
        \includegraphics[page=3, width=0.22\linewidth]{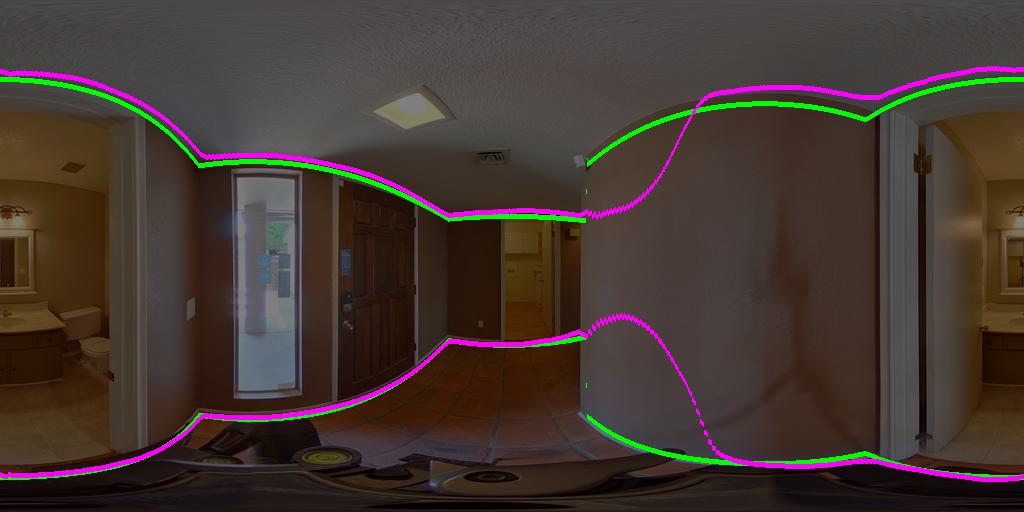}

        \includegraphics[page=3, width=0.22\linewidth]{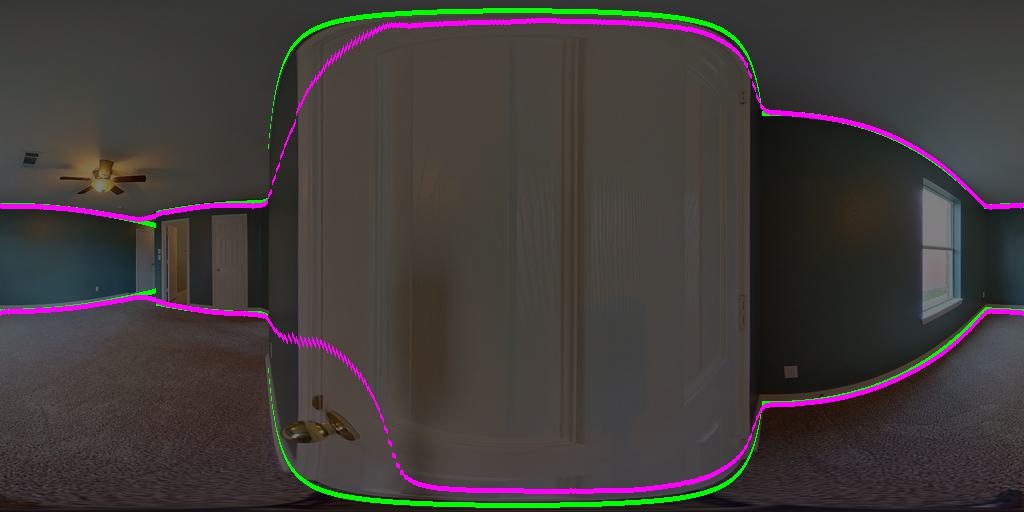}
        \\
        \includegraphics[page=11, width=0.22\linewidth]{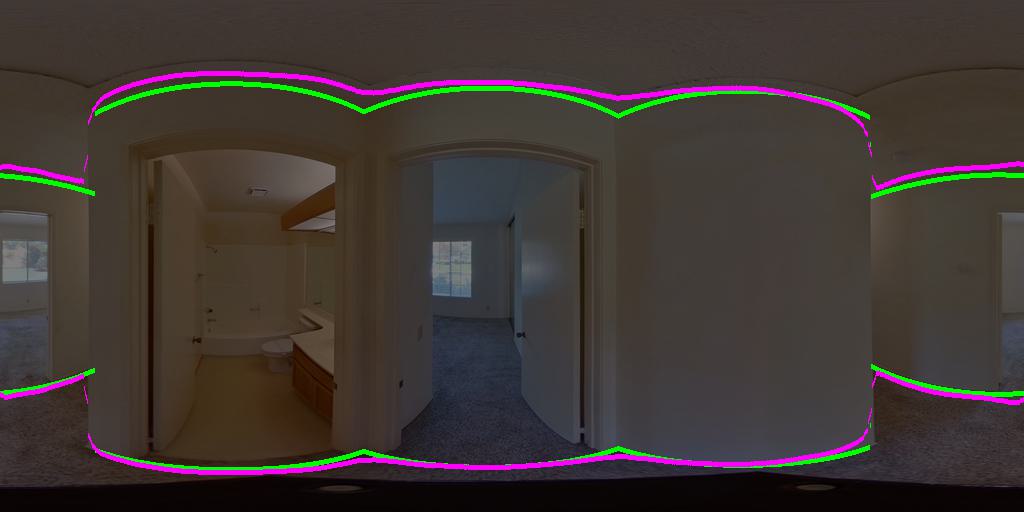}

        \includegraphics[page=13, width=0.22\linewidth]{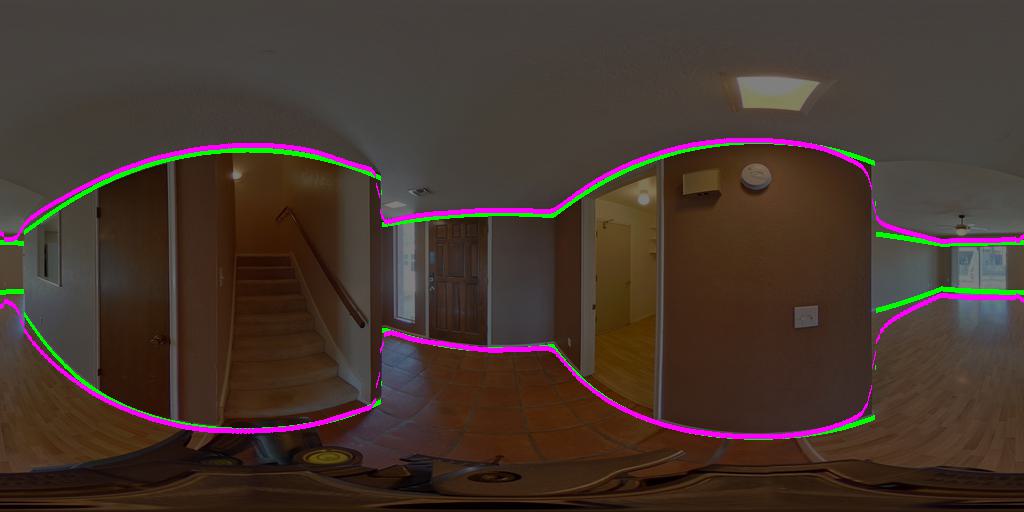}
        
        \includegraphics[page=3, width=0.22\linewidth]{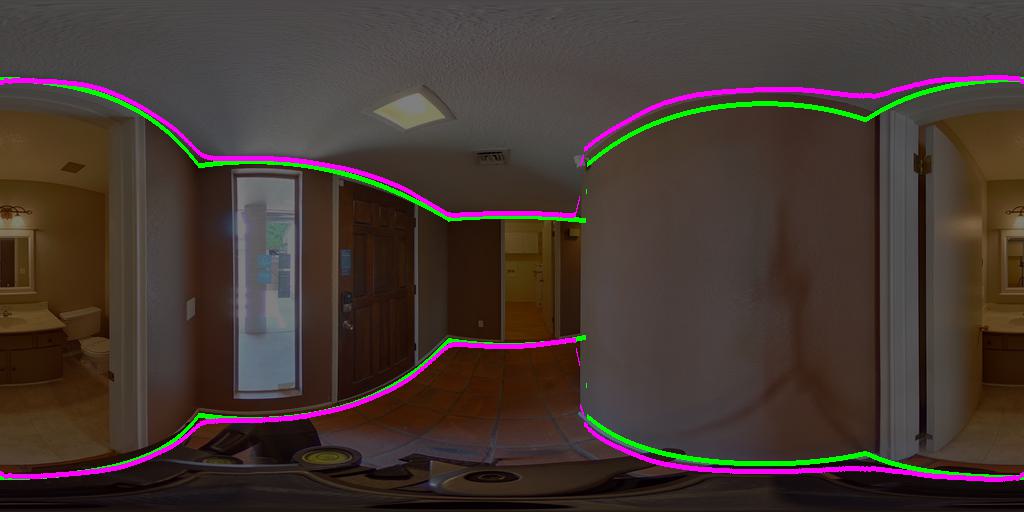}

        \includegraphics[page=3, width=0.22\linewidth]{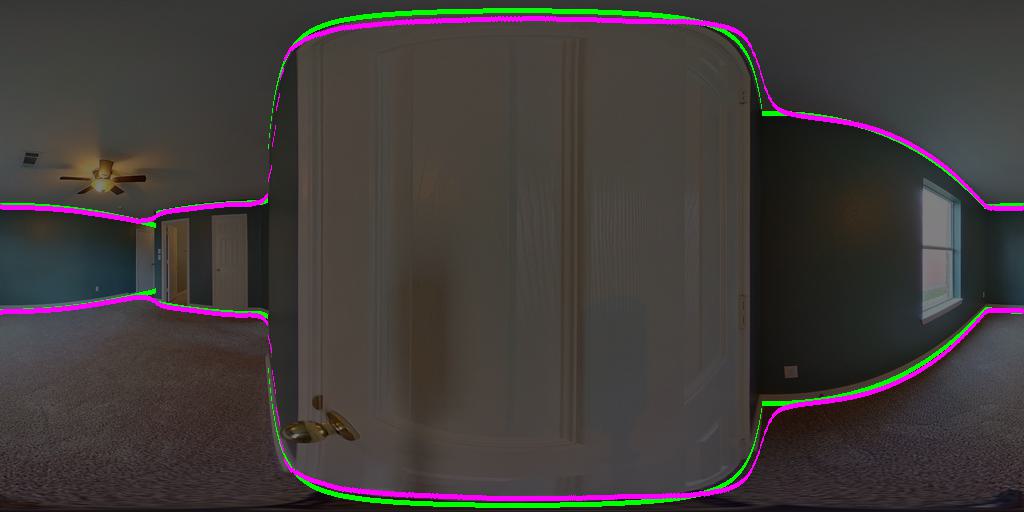}
        \\
        \\

    \end{tabular}
    
    \begin{picture}(0, 0)
    \put(-167,232){\begin{turn}{90} 
    Ours
    \end{turn}
    }
    \put(-167,127){\begin{turn}{90} 
    Ours
    \end{turn}
    }
    \put(-167,22){\begin{turn}{90} 
    Ours
    \end{turn}
    }

    \put(-167,265){\begin{turn}{90} 
    360-MLC
    \end{turn}
    }
    \put(-167,160){\begin{turn}{90} 
    360-MLC
    \end{turn}
    }
    \put(-167,55){\begin{turn}{90} 
    360-MLC
    \end{turn}
    }

    \put(-100,210){
    Qualitative comparisons on our HM3D-MVL dataset
    }

    \put(-90,105){
    Qualitative comparisons on MP3D-FPE~\cite{360_dfpe}
    }

    \put(-80,0){
    Qualitative comparisons on ZInD~\cite{zind}
    }
    \end{picture}
    \caption{
        \textbf{Qualitative comparisons on panoramic images.}. We present the results of room layout estimation after self-training using 360-MLC~\cite{360_mlc} and our proposed framework. Results are evaluated in three different datasets: 1) at the top on our proposed HM3D-MVL, 2) in the middle on MP3D-FPE~\cite{360_dfpe}, and 3) at the bottom on the real-world dataset ZInD~\cite{zind}. The green lines represent the ground truth reference and the magenta lines represent the layout estimations.
        % In column (a) our proposed dataset \textbf{HM3D-MVL}, column (b) MP3D-FPE \cite{360_dfpe}, and column (c) ZInD \cite{zind}.  
    }
    \label{fig_qualitative_pano}
\end{figure*}

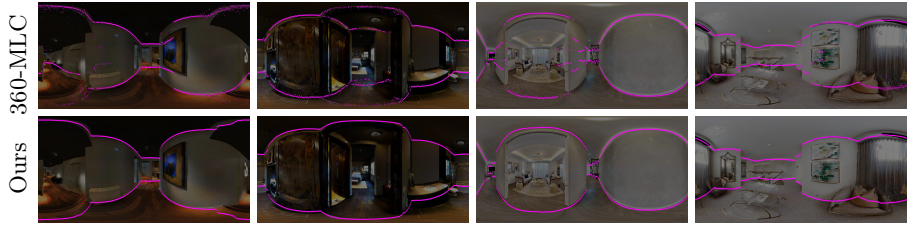
\begin{figure*}
    \small
    \centering
    \footnotesize
    \begin{tabular}{c c c c}
        % ############################## HM3D-mvl
        \includegraphics[page=9, width=0.23\linewidth]{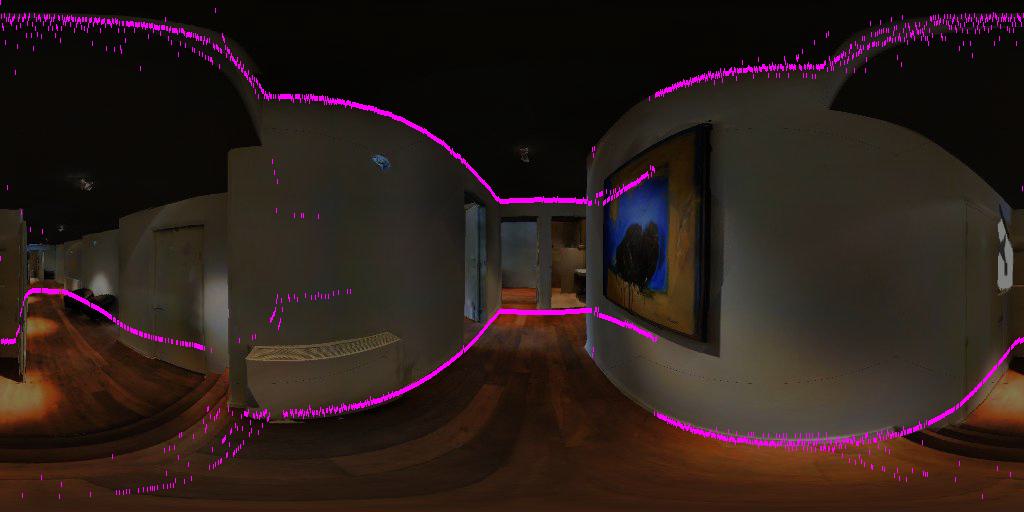}

        \includegraphics[page=9, width=0.23\linewidth]{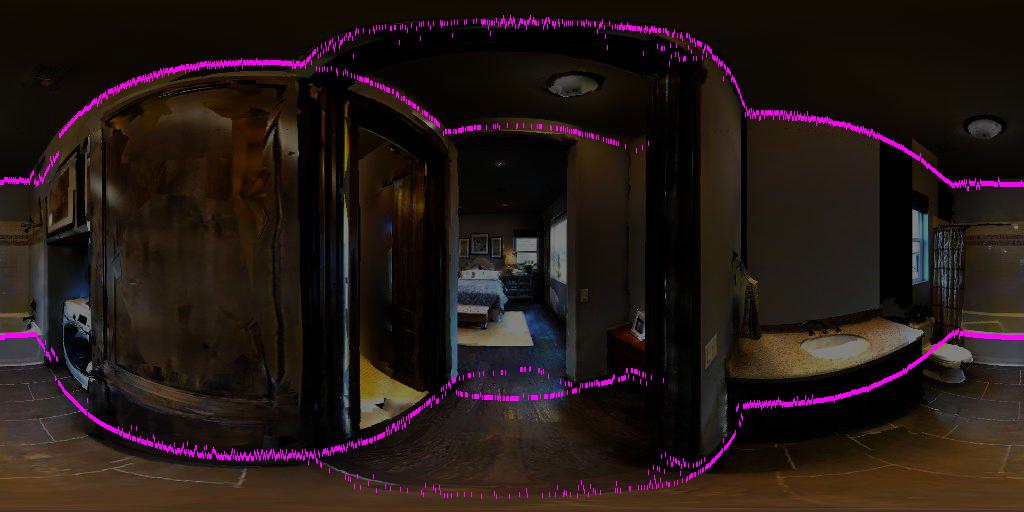}

        \includegraphics[page=9, width=0.23\linewidth]{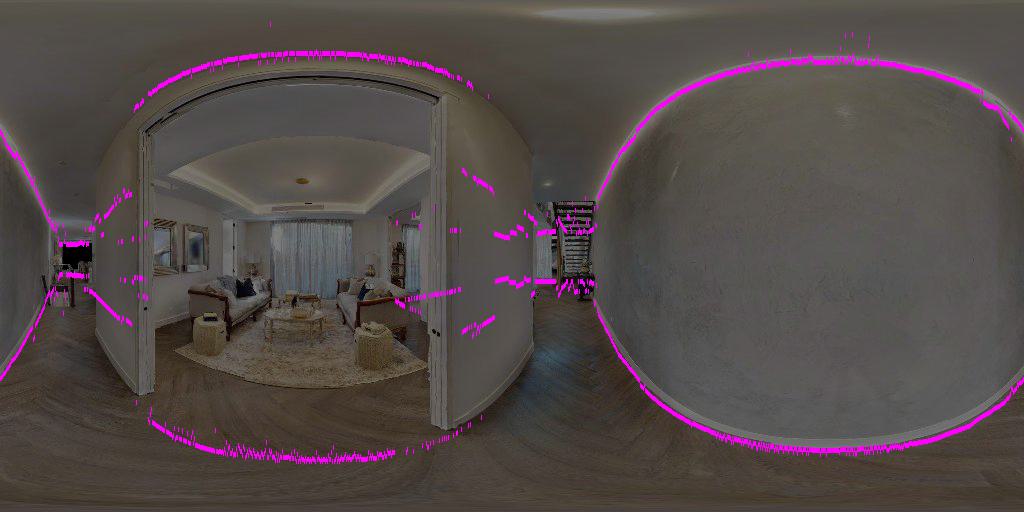}

        \includegraphics[page=9, width=0.23\linewidth]{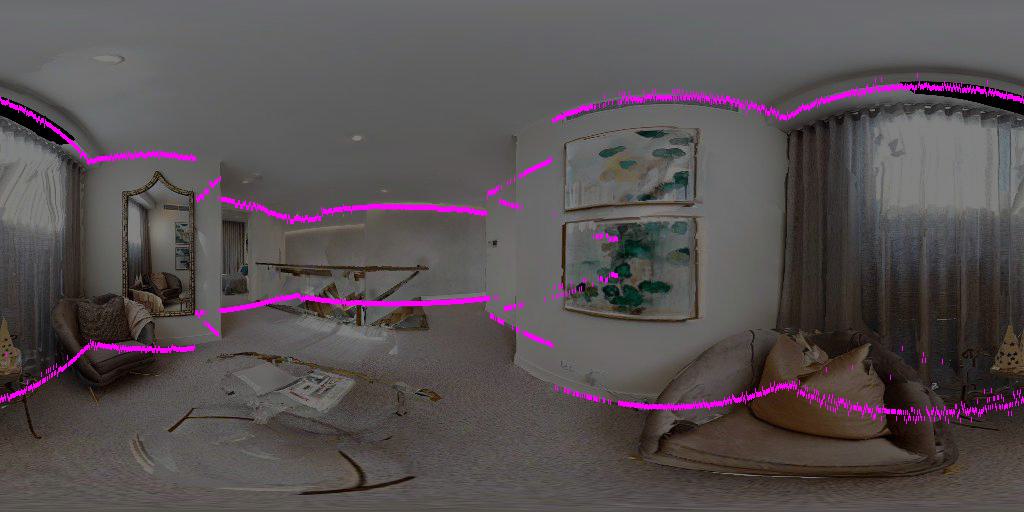}
        \\
        \includegraphics[page=9, width=0.23\linewidth]{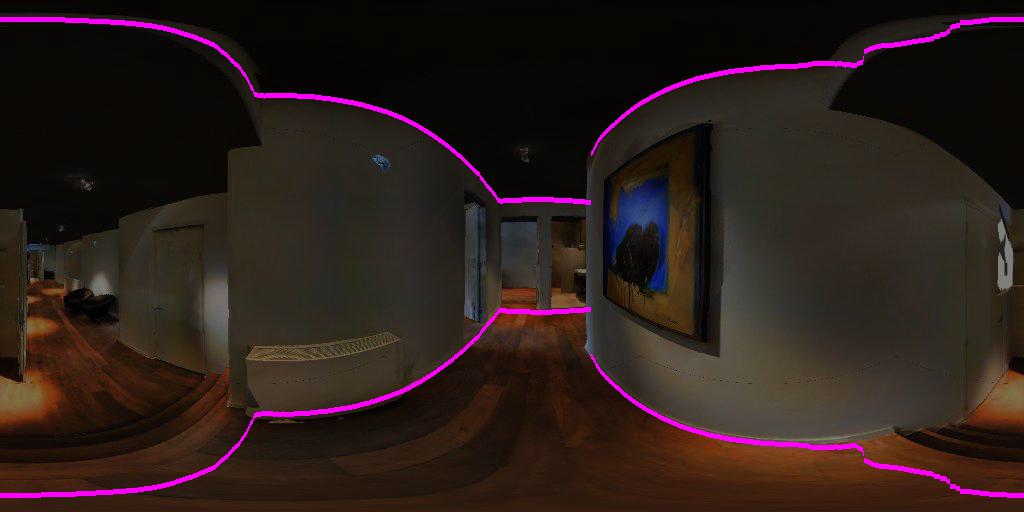}

        \includegraphics[page=9, width=0.23\linewidth]{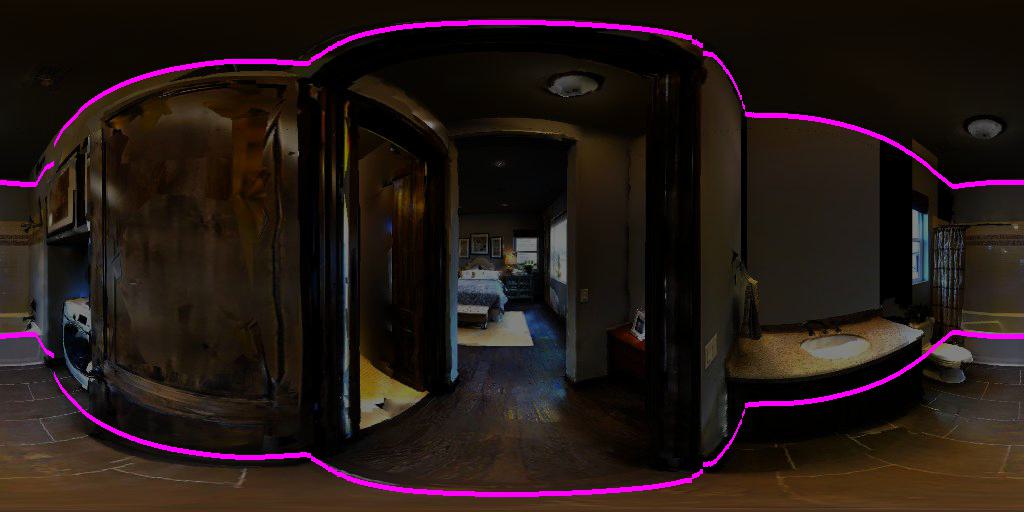}

        \includegraphics[page=9, width=0.23\linewidth]{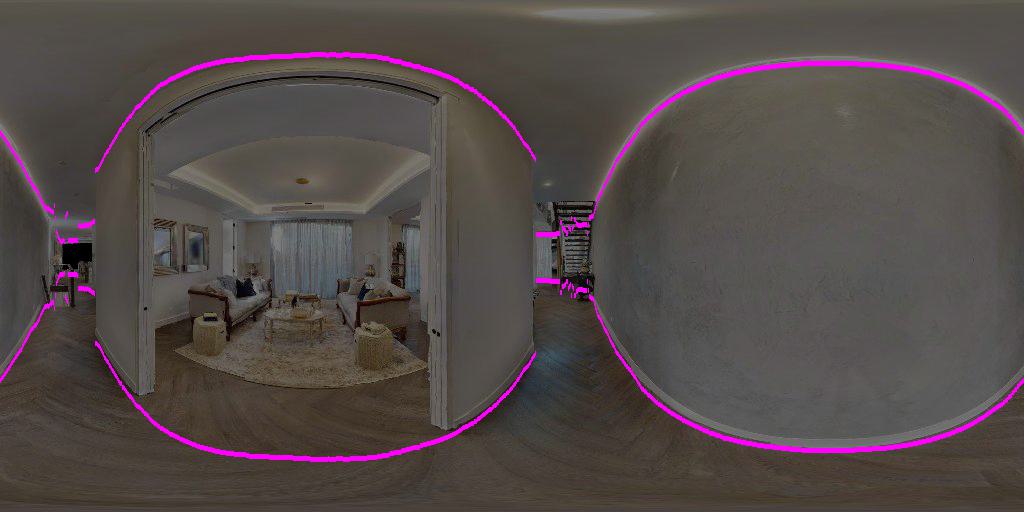}

        \includegraphics[page=9, width=0.23\linewidth]{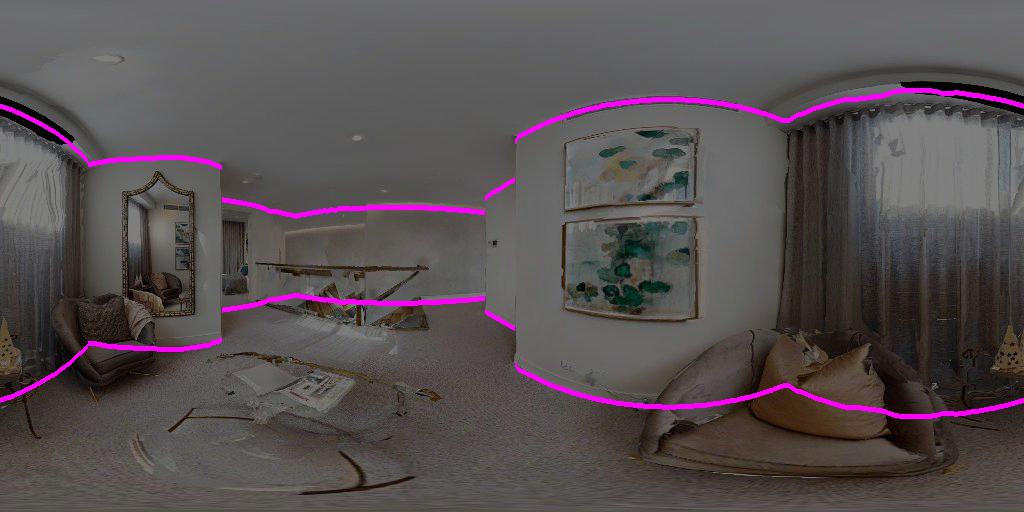}

        \\
    \end{tabular}

    \begin{picture}(0, 0)

    \put(-175,45){\begin{turn}{90} 
    360-MLC
    \end{turn}}
  
    \put(-175,15){\begin{turn}{90} 
    Ours
    \end{turn}}

    \end{picture}
    
    \caption{
        \textbf{Qualitative comparisons of pseudo labels on panoramic images.} We present the qualitative results of estimated pseudo labels (magenta lines) on the panoramic images: 1) the first row, 360-MLC~\cite{360_mlc}; 2) the second row, our ray-casting pseudo labels. 
        % \todo{Add detail of occlusion and entrance handling.}
    }
    \label{fig_qualitative_labels_equi}
\end{figure*}

\kike{In this section, we present qualitative results for our proposed ray-casting pseudo-labeling framework. These results are presented in \cref{fig_qualitative_labels_equi} and \cref{fig_qualitative_bev}, where the former presents pseudo-labels projected on panoramic images and the latter presents pseudo-labels projected in BEV. Based on the results in~\cref{fig_qualitative_labels_equi}, we corroborate our hypothesis that our ray-casting 
pseudo-labels can handle occluded geometries better than 360-MLC. Furthermore, we find evidence that challenging views such as entrance and gates are better defined by our proposed pseudo-labels. This evidence aligns with our findings in \cref{fig_qualitative_pano}, where results of a self-trained model using our proposed framework show better estimation for such challenging view locations.}
\kike{Furthermore, based on the results presented in \cref{fig_qualitative_bev}, we can assert that our ray-casting pseudo-labels yield a less noisy geometry compared to 360-MLC, as well as it is capable of defining circular walls directly from multiple estimations.}
\vspace{-3mm}
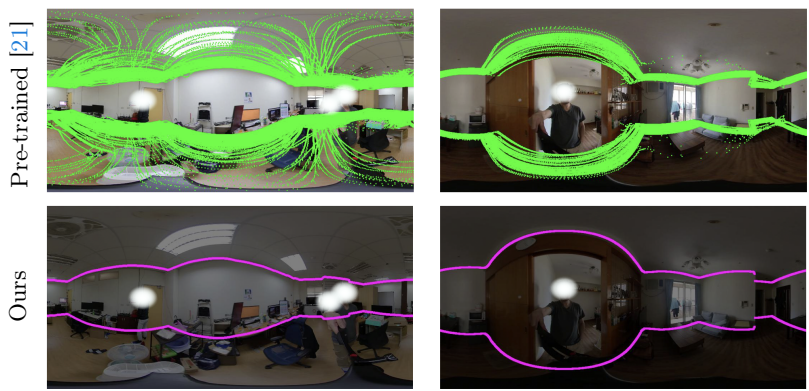
\begin{figure}[t!]
\centering
\includegraphics[width=\linewidth]
{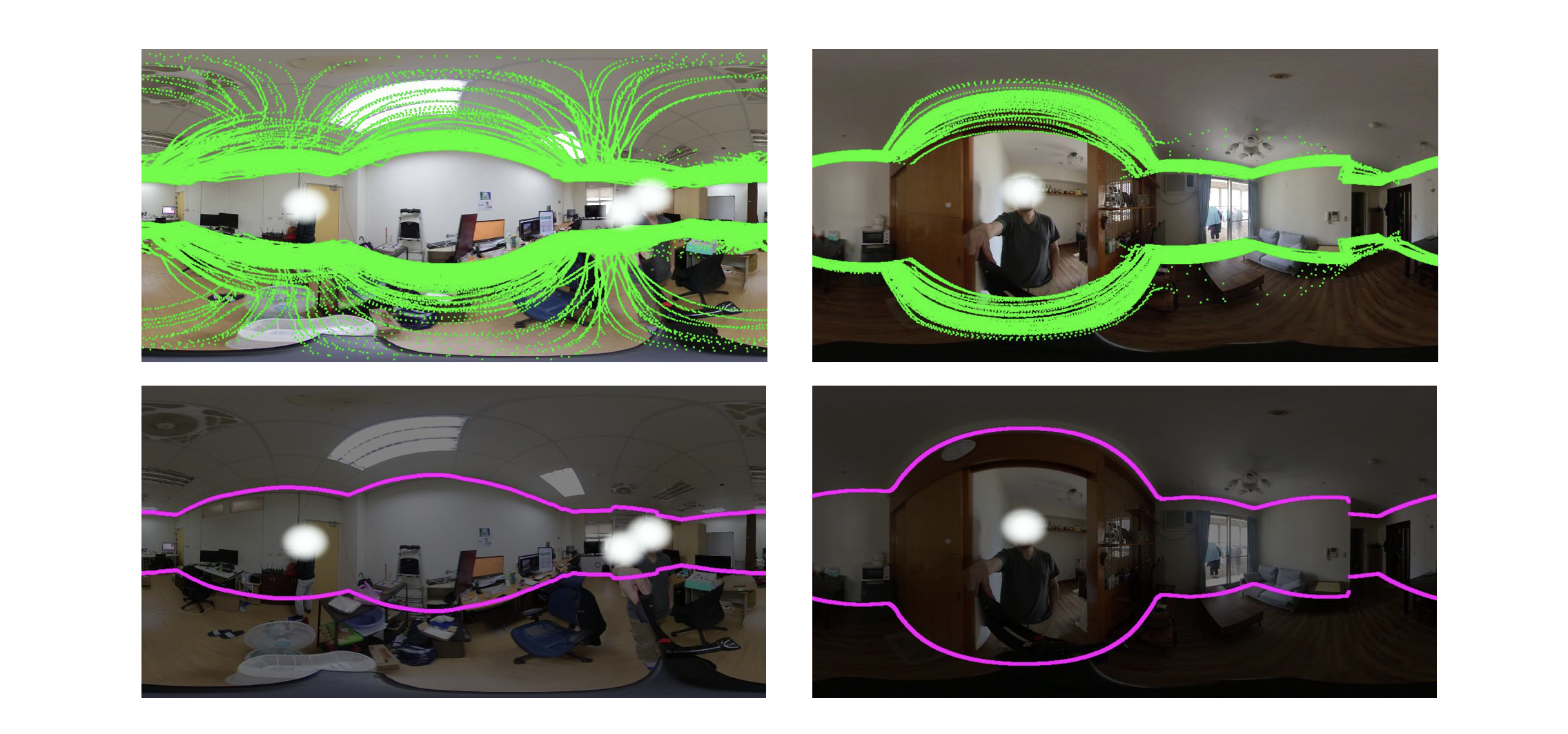}
    \begin{picture}(0, 0)
    \put(-155,95){\begin{turn}{90} 
    Pre-trained~\cite{horizonnet}
    \end{turn}
    }
    \put(-155,44){\begin{turn}{90} 
    Ours
    \end{turn}
    }
    \end{picture}
    \vspace{-5mm}
    \caption{
        \textbf{Qualitative results in real-world scenes.} We show layout boundaries estimated in real-world data using a hand-handled camera (Insta360). In the first row, we illustrate all layouts estimated from a pre-trained model~\cite{horizonnet}. In the second row, we show the results of our ray-casting pseudo labeling process presented in~\cref{sec:ray_casting}.}
    \label{fig_real_world}
    \vspace{-6mm}
\end{figure}

\vspace{-2mm}
\subsubsection{Qualitative Results on Real-world Data.}
\kike{    }\kike{In~\cref{fig_real_world}, we present two qualitative results in two real-world scenes, demonstrating the versatility of our ray-casting pseudo-labeling in real-world scenarios. For these experiments, we collect several panoramic images using a commercial camera, Insta360\footnote{https://www.insta360.com/}, and estimate their camera poses using OpenVSLAM~\cite{openvslam2019}. Subsequently, we register each image with its corresponding layout estimation (utilizing HorizonNet~\cite{horizonnet} pre-trained in~\cite{mp3d}) by using the layout registration method outlined in~\cite{360_dfpe}. In the first row, we present evidence of the domain gap in the pre-trained model showing a significant level of noise in the boundary layout estimations for both depicted scenes. In the second row, we present the results of our proposed ray-casting pseudo-labeling framework presented in \cref{sec:ray_casting}. Note that our solution is capable of aggregating multiple noisy estimates to define a reliable underlying geometry for self-training remarkably.}
\vspace{-3mm}
\subsection{Ablation Study for Weighted Distance Loss Formulation}
\begin{table}[!t]
\vspace{5mm}
\fontsize{8.5}{10}\selectfont
  \caption{Ablation study for our weighted-distance loss using 10\% of data.
  % losses. PT stands for Pre-Trained. WD stands for Weighting Distance (Our full MLC++).
  }
  \vspace{-2mm}
  \label{tab_ablation}
  \centering
  \begin{tabular}{ll c c | c c}
    \toprule
     && \multicolumn{2}{c|}{Testing set} & \multicolumn{2}{c}{Occlusion Subset}\\
     & Loss  & 2D IoU~$\uparrow$ & 3D IoU~$\uparrow$ & 2D Io~$\uparrow$ & 3D IoU~$\uparrow$ \\
    \midrule
    (a)& Pre-trained~\cite{horizonnet} &76.71	&71.79	&78.74&	75.72\\ 
    (b)& Pseudo-labels &81.65& 76.99 &80.85	&78.98\\
    (c)& $\omega$=$\sigma^{-2}$ &
    81.02	&76.58	&81.28	&79.53\\
    % (d)& $\omega=sin(\sigma^{-2}_{bev})$.& 
    % 81.01	&77.31	&81.97	&80.19\\
    (d)& $\omega$ = \cref{eq_w_fucntion}  
    &\textbf{81.74}	&\textbf{77.99}	
    &\textbf{82.05}	&\textbf{80.45}\\ 
    \bottomrule
  \end{tabular}
  \vspace{-5mm}
\end{table}
%\cref{eq_w_fucntion}
% \vspace{-3mm}

\kike{We present an ablation study that validates our weighted distance loss formulation presented in \cref{sec:w_fucntion}. The results of this ablation are shown in \cref{tab_ablation}. By comparing rows (a) and (b), we validate the gain in performance of self-training directly using our proposed ray-casting pseudo-labels without any weighting formulation. By comparing (c) and (b), we verify a weighted formulation based only on the uncertainty $\sigma$ computed by \cref{eq_ray_cast_min}. We can appreciate that this weighting formulation yields better performance on the occlusion subset but not for the whole testing set. We argue that a weighting formulation based on uncertainty $\sigma$ does not consider any geometry information. In contrast, in row (d), we show the results of our weighted formulation as presented in \cref{eq_w_fucntion}. Thus we can assert that a weighting formulation that prioritizes the farthest geometries with respect to the camera view yields better performance.
}

\vspace{-3mm}
\section{Conclusions}
\label{sec:conclusion}
% \vspace{-2mm}
\kike{In this paper, we present a geometry-aware self-training framework for multi-view room layout estimation that requires only unlabeled images as input. Our approach utilizes a ray-casting formulation capable of handling occluded geometries directly from noisy estimations. To further exploit the benefit of the multi-view setting, we propose a weighted distance loss function that focuses on the farthest geometries in the scene. Through a comprehensive evaluation using different datasets, room layout models, and settings, we demonstrate the state-of-the-art performance of our solution.}

\section*{Acknowledgements}
This project is supported by The National Science and Technology Council NSTC and The Taiwan Computing Cloud TWCC under the project NSTC 112-2634-F-002-006.

% \bibliography{main}

\end{document}